\documentclass[]{stresearch}

\usepackage{hyperref}
\usepackage{cleveref}
\usepackage{amssymb}
\usepackage{verbatim}

\usepackage{wrapfig}
\usepackage{graphicx}
\usepackage{floatrow}
\usepackage{placeins}
\usepackage{subcaption}
\usepackage{listings}
\usepackage{algorithm}
\usepackage{subcaption}
\usepackage{dsfont}
\usepackage{enumitem}
\usepackage{float}
\usepackage{mmstyles}
\usepackage[toc,page,header]{appendix}

\usepackage{xspace}
\usepackage[utf8]{inputenc}  
\usepackage[T1]{fontenc}     
\usepackage{CJKutf8} 

\usepackage[misc]{ifsym}

\usepackage[toc,page,header]{appendix}
\newcommand{\methodname}{SenseNova-MARS}
\newcommand{\datasetname}{HR-MMSearch}

\title{SenseNova-MARS: Empowering Multimodal Agentic Reasoning and Search
via Reinforcement Learning}

\author{
Yong Xien Chng$^{*,1,2}$, 
Tao Hu$^{*,1,3}$, 
Wenwen Tong$^{*,\dagger, 1}$, 
Xueheng Li$^{1,3}$, 
Jiandong Chen$^{1}$,
\\
Haojia Yu$^{1}$, 
Jiefan Lu$^{1}$, 
Hewei Guo$^{1}$, 
Hanming Deng$^{1}$, 
Chengjun Xie$^{3}$,
\\
Gao Huang$^{2}$,  
Dahua Lin$^{1}$, 
Lewei Lu$^{\textrm{\Letter},1}$
\\[4px] 
\parbox{\textwidth}{\centering\normalsize
    $*$ Equal Contribution, \,
    $\dagger$ Project Lead, \,\,
    $\textrm{\Letter}$ Corresponding Author \\[4px] 
    $^1$SenseTime Research, \,
    $^2$Tsinghua University, \,
    $^3$University of Science and Technology of China
}}

\abstract{
While Vision-Language Models (VLMs) can solve complex tasks through agentic reasoning, their capabilities remain largely constrained to text-oriented chain-of-thought or isolated tool invocation. They fail to exhibit the human-like proficiency required to seamlessly interleave dynamic tool manipulation with continuous reasoning, particularly in knowledge-intensive and visually complex scenarios that demand coordinated external tools such as search and image cropping.
In this work, we introduce SenseNova-MARS, a novel Multimodal Agentic Reasoning and Search framework that empowers VLMs with interleaved visual reasoning and tool-use capabilities via reinforcement learning (RL).
Specifically, SenseNova-MARS dynamically integrates the image search, text search, and image crop tools to tackle fine-grained and knowledge-intensive visual understanding challenges. 
In the RL stage, we propose the Batch-Normalized Group Sequence Policy Optimization (BN-GSPO) algorithm to improve the training stability and advance the model’s ability to invoke tools and reason effectively.
To comprehensively evaluate the agentic VLMs on complex visual tasks, we introduce the HR-MMSearch benchmark, the first search-oriented benchmark composed of high-resolution images with knowledge-intensive and search-driven questions. 
Experiments demonstrate that SenseNova-MARS achieves state-of-the-art performance on open-source search and fine-grained image understanding benchmarks.
Specifically, on search-oriented benchmarks, SenseNova-MARS-32B scores 74.3 on MMSearch and 54.4 on HR-MMSearch, surpassing proprietary models such as Gemini-3-Pro and GPT-5.2.
SenseNova-MARS represents a promising step toward agentic VLMs by providing effective and robust tool-use capabilities. To facilitate further research in this field, we will release all code, models, and datasets.

\date{\today}
\checkdata[Codebase]{\url{https://github.com/OpenSenseNova/SenseNova-MARS}}
\checkdata[Model]{\url{https://huggingface.co/sensenova/SenseNova-MARS-8B}}
}

\begin{document}
\maketitle

\section{Introduction}\label{sec:intro}

\begin{figure}[t]
    \centering
    \includegraphics[width=0.9\textwidth]{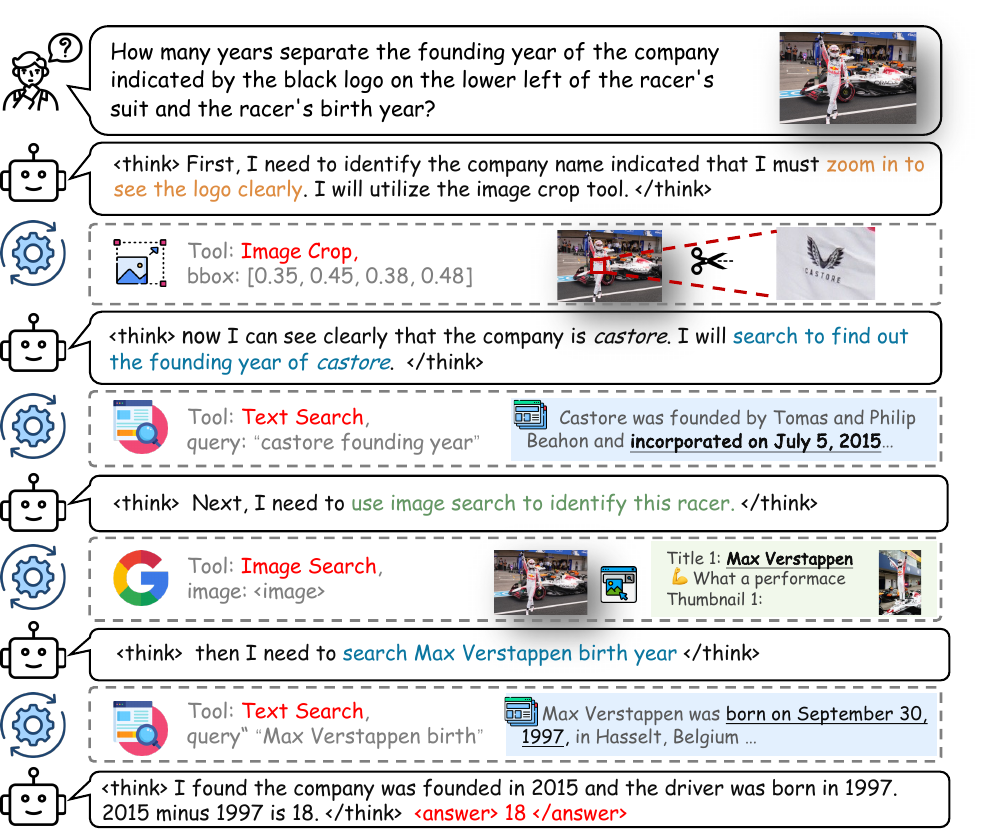}
     \caption{Reasoning trajectory of \methodname{}. \methodname{} tackles the challenging visual task by leveraging an integrated suite of text search, image search, and image crop tools within the reasoning process.}
    \label{fig:teaser}
\end{figure}

Vision-Language Models (VLMs) have significantly advanced the development of artificial general intelligence with impressive capabilities across a wide range of visual understanding tasks \cite{bai2025qwen2, chen2024far,wang2025internvl3, hurst2024gpt, comanici2025gemini}.
To tackle more complex real-world challenges, recent efforts have equipped VLMs with reasoning capabilities and external tools, such as search engines for knowledge-intensive tasks or cropping tools for fine-grained visual analysis \cite{wu2025mmsearch, guo2025deepseek, hong2025deepeyesv2, Qwen3-VL}.
However, these systems remain constrained by text-centric reasoning chains and isolated tool calls, lacking the dynamic integration of multimodal reasoning and tool use.
As illustrated in Fig.~\ref{fig:teaser}, real-world visual understanding demands agentic models that can interleave planning, reasoning, and multi-tool execution into a cohesive, adaptive, and multi-step process.
Therefore, it is critical to develop agentic models with robust tool invocation and human-like multi-step reasoning capabilities to solve challenging visual tasks.

\begin{figure}[t]
    \centering
    \includegraphics[width=\textwidth]{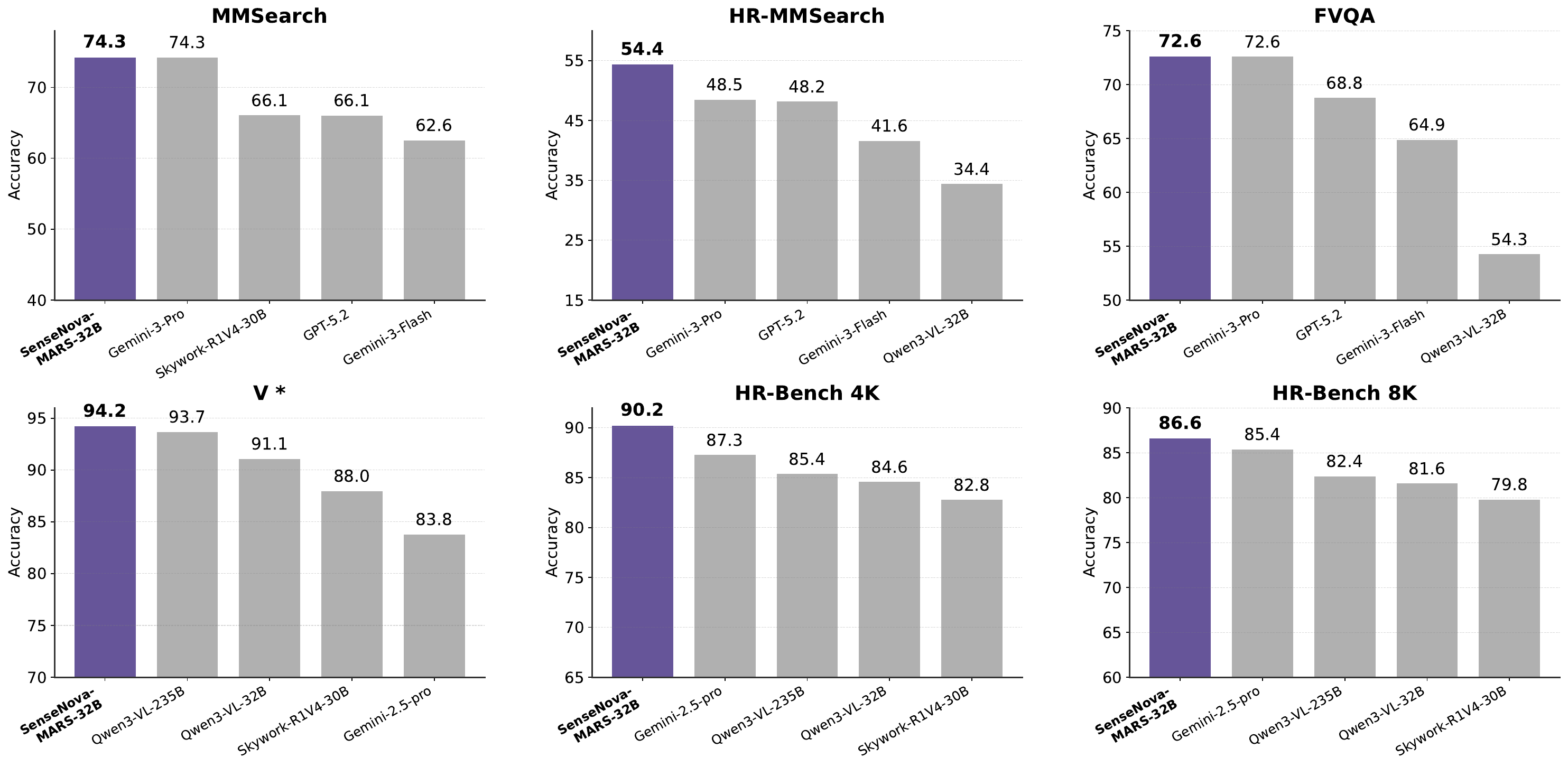}
     \caption{Overall performance of \methodname{}-32B compares to other models across six benchmarks. \methodname{}-32B demonstrates exceptional performance on the search-oriented benchmarks such as MMSearch \cite{jiang2024mmsearch}, HR-MMSearch and FVQA \cite{wu2025mmsearch}, surpassing leading proprietary models such as Gemini-3-Pro \cite{gemini3pro} and GPT-5.2 \cite{openai2025gpt5_2}.
For the high-resolution perception benchmark such as V* Bench~\cite{wu2024v} and 
HR-Bench \cite{wang2025divide}, \methodname{}-32B can outperform top-tier models such as Qwen3-VL-235B-A22B \cite{Qwen3-VL}.}
 \label{fig:overall_performance}
\end{figure}

The potential of end-to-end reinforcement learning (RL) with Group Relative Policy Optimization (GRPO) to enhance the reasoning capabilities has been highlighted by models such as DeepSeek-R1 \cite{guo2025deepseek}, establishing RL as the mainstream approach for developing reasoning models.
Existing search-based agents \cite{li2025search, jin2025search,dong2025tool} have utilized end-to-end RL to enhance model performance through multi-turn reasoning with text search tools, demonstrating substantial performance gains over Retrieval-Augmented Generation (RAG) methods.
MMSearch-R1 \cite{wu2025mmsearch} further integrates image search with text search tools, enabling VLM to interact with the real-world environment to tackle knowledge-intensive and info-seeking visual tasks.
However, it remains inadequate for high-resolution perception tasks as shown in Fig.~\ref{fig:teaser}, which requires a level of detailed visual reasoning and fine-grained analysis that cannot be achieved with image and text search tools alone.
Recent advances such as OpenAI-o3 \cite{OpenAI2025Thinking} introduced the ``\textit{Thinking with images}'' paradigm, which interleaves image and text reasoning by employing the image crop tool to perform fine-grained analysis of complex visual scenes \cite{su2025thinking}.
Subsequent work, including Pixel Reasoner \cite{su2025pixel} and DeepEyes \cite{zheng2025deepeyes}, further demonstrates that pixel-space visual reasoning capabilities can be incentivized by RL, enabling VLMs to learn how to interact with image manipulation tools and proactively gather necessary visual information. 
These approaches are typically restricted to either search-based or image manipulation tools, which hinders their ability to handle dynamic real-world visual tasks requiring both. 
This limitation necessitates the creation of search-oriented agentic VLMs that can leverage tools like image crop to handle complex visual scenes, driven by end-to-end RL.


To this end, we propose \methodname{}, a novel Multimodal Agentic Reasoning and Search framework that leverages RL to integrate image search, text search, and image crop tools into a dynamic, multi-turn reasoning process.
Specifically, \methodname{} adaptively interacts with the tool sets through reasoning and planning, learning when and how to invoke the search and image crop tools during the iterative reasoning process.
We systematically investigate how to build this agentic VLM based on Qwen2.5-VL \cite{bai2025qwen2}, and Qwen3-VL \cite{Qwen3-VL}, introducing \methodname{} in three scales including
\methodname{}-7B, \methodname{}-8B and \methodname{}-32B.
Our approach focuses on three key aspects: training data construction, a two-stage training pipeline including the cold-start stage and RL, and the design of RL algorithm.
We first curate the high-quality cold-start and RL training data through a synthesis pipeline with rigorous quality verification.  
The dataset spans diverse visual tasks, including knowledge-intensive, high-resolution perception, and search-oriented tasks, which necessitate the use of the search tool, image crop tool, or a combination of both.
During the initial cold-start supervised fine-tuning (SFT) stage, the model learns basic tool-usage patterns from a minimal dataset of approximately 3,000 samples.
This compact dataset is crucial as it establishes a foundation for the subsequent RL stage, particularly in learning to use the tools effectively.
To advance the agentic VLM's capabilities in multi-tool collaboration and reasoning during the RL stage, we propose Batch-Normalized Group Sequence Policy Optimization (BN-GSPO), an extension of the standard GSPO algorithm.
Experiments indicate that BN-GSPO can improve training stability for multi-tool rollout trajectories and yield significant performance gains.

To further evaluate visual search agents for complex scenes, we introduce \datasetname{}, the first search-oriented benchmark composed of high-resolution images with knowledge-intensive and search-driven questions. 
\methodname{} demonstrates superior performance on \datasetname{}, far beyond the search-only or crop-only models, and is comparable with proprietary models like Gemini-3-Pro \cite{gemini3pro}, and GPT-5.2 \cite{openai2025gpt5_2}.
Furthermore, \methodname{}-32B achieves state-of-the-art (SOTA) results across diverse open-source search-oriented and perception benchmarks as shown in Fig.~\ref{fig:overall_performance}, attaining a leading score of 74.3 in MMSearch \cite{jiang2024mmsearch} and 54.4 in HR-MMSearch.
For the high-resolution perception benchmark, \methodname{}-32B achieves 94.2 on V* Bench~\cite{wu2024v} and 90.2 on 
HR-Bench 4K \cite{wang2025divide}, exceeding top-tier models like Qwen3-VL-235B-A22B \cite{Qwen3-VL} and  Skywork-R1V4-30B \cite{zhang2025skywork}.
Similarly, our smaller variant, \methodname{}-8B, is superior to Gemini-3-Flash \cite{gemini3flash} and GPT-5 \cite{openai2025gpt5}, and achieves SOTA among similarly-sized open-source models such as MMSearch-R1 \cite{wu2025mmsearch} and DeepEyesV2 \cite{hong2025deepeyesv2}.

To sum up, our contributions are threefold: (1) We propose \methodname{}, the first end-to-end agentic high-resolution vision-language model developed by RL, with the capability of performing image search, text search, and image crop tools within the reasoning process.
(2) We introduce \datasetname{}, the first benchmark specifically designed for high-resolution, search-oriented, and knowledge-intensive tasks, enabling comprehensive evaluation of agentic reasoning and multi-tool invocation capabilities in VLMs.
(3) We propose BN-GSPO, a stable and high-performing RL algorithm for training agentic VLMs in tool invocation. Extensive experiments on diverse open-source benchmarks with SOTA performance demonstrate that \methodname{} exhibits strong reasoning and robust tool invocation, validating the effectiveness of our approach.


\section{Related Works}\label{sec:related_works}
\subsection{Search-based Agentic VLMs}
To mitigate the restriction of static knowledge bases, large language models have evolved from static retrieval mechanisms to dynamic, tool-augmented reasoning.
Early RAG systems such as Dense Passage Retrieval~\cite{karpukhin2020dense} and Retrieval-Augmented Generation~\cite{lewis2020retrieval} rely on static document corpora for text grounding. 
Later, search-augmented systems such as WebGPT~\cite{nakano2021webgpt}, Toolformer~\cite{schick2023toolformer}, and SAIL~\cite{luo2023sail} allow models to access live web information and reason over up-to-date content. 
In parallel, multimodal models including 
REVEAL~\cite{hu2023reveal}, 
RagVL~\cite{chen2024mllm}, 
and VisRAG~\cite{yu2024visrag} show that combining textual and visual retrieval improves knowledge-intensive understanding. 
However, these RAG-based methods follow the fixed workflow and often result in excessive retrieval, which is suboptimal in practice.
Recently, text-based agentic search models such as Search R1~\cite{jin2025search} and Search o1~\cite{li2025search} integrate search tools into the chain of thought to enhance performance on knowledge-intensive tasks. 
In the multimodal domain, agentic search models such as MMSearch-R1~\cite{wu2025mmsearch} train VLMs to trigger image and text searches dynamically, while WebWatcher~\cite{geng2025webwatcher} extends this approach through synthetic trajectories that improve generalization. 
Existing agentic VLM research has mainly focused on holistic image understanding that gathers broad contextual information while overlooking the analysis of fine-grained image regions. 
This limitation reduces their ability to answer region-specific questions and to perform tasks that require precise visual grounding.


\subsection{Thinking with Images}
The ``\textit{Thinking with images}'' paradigm, introduced by OpenAI-o3~\cite{OpenAI2025Thinking}, has spurred the development of agentic visual reasoning VLMs that can interleave image and text reasoning with iterative visual analysis~\cite{su2025thinking, su2025openthinkimg}. DeepEyes~\cite{zheng2025deepeyes} provides an open-source implementation, demonstrating that end-to-end RL can incentivize models to adopt this behavior, significantly improving performance on fine-grained visual tasks. 
However, subsequent works reveal that pure RL training is insufficient for complex, multi-turn interaction~\cite{su2025pixel, zhang2025thyme, lai2025mini, zhou2025reinforced}. 
Pixel Reasoner~\cite{su2025pixel} identifies a critical "learning trap" where models bypass nascent visual tools. 
To address this, it proposes a two-phase approach: a cold-start phase to first establish foundational tool use, followed by a curiosity-driven RL phase to incentivize pixel space exploration. 
Similarly, Mini o3~\cite{lai2025mini} observes that pure RL cannot generate the deep trajectories for hard visual searches and therefore adopted a two-stage training method to activate multi-turn capabilities. Despite these advances, current methods remain fundamentally knowledge-limited, as their toolsets focus solely on perceptual image operations and are inherently ill-equipped for tasks that demand open-web access or external knowledge.

\subsection{Multi-tool Agentic VLMs}
Recent advancements in VLMs \cite{bai2025qwen2, xu2025qwen2, wang2025internvl3, hurst2024gpt}, particularly the integration of RL paradigms, have catalyzed the rapid evolution of sophisticated agentic systems \cite{dong2025agentic, wang2025mllm, jiang2025verltool}. These models demonstrate impressive capabilities in complex problem-solving through hierarchical planning, reasoning, and the strategic invocation of diverse external tools (e.g., image processing, code execution, web search). 
Notably, recent studies \cite{li2025torl, feng2025retool, singh2025agentic} have highlighted the crucial role of RL in enabling agents to develop adaptive multi-tool utilization strategies, optimizing both tool-calling policies and reasoning trajectories through interaction with the environment and feedback from execution outcomes. 
Visual-ARFT~\cite{liu2025visual} introduces a  framework that empowers agents to perform multi-tool collaboration by integrating web information retrieval with autonomous code execution for visual reasoning. 
Similarly, DeepMMSearch-R1~\cite{narayan2025deepmmsearch} proposes a VLM-based agent equipped with advanced tool integration for web search, orchestrating textual queries alongside the image search mechanism that leverages intermediate cropping to concentrate on relevant visual entities. However, existing methods struggle to effectively coordinate the acquisition of critical visual information with external knowledge retrieval, limiting their capability for collaborative tool invocation and coherent reasoning.

\begin{figure*}
\vspace{-0.25cm}
    \centering
    \includegraphics[width=0.95\textwidth]{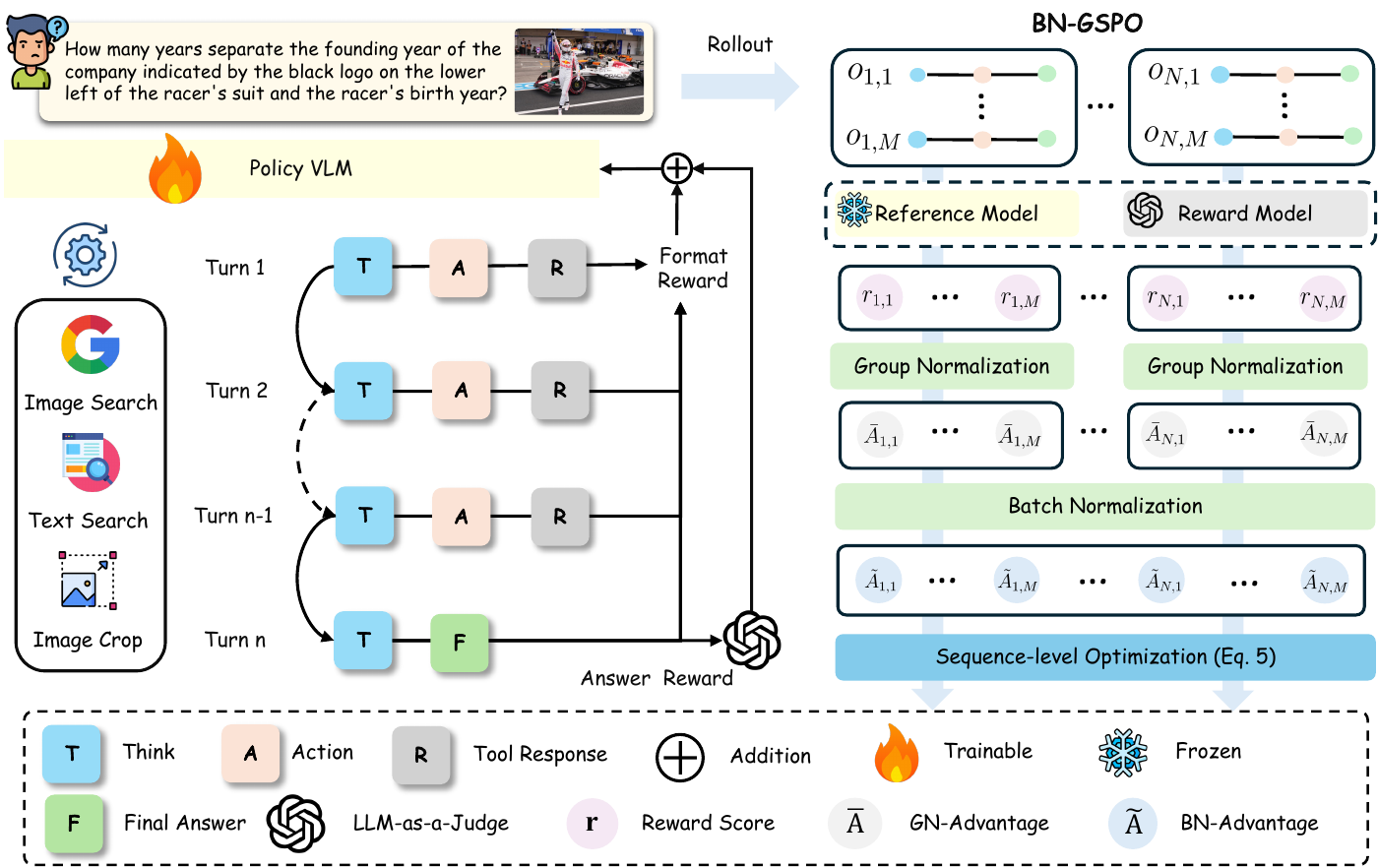}
    \caption{The illustration of \methodname{} RL training pipeline. \methodname{} adaptively invokes the image search, text search and image crop  tools in the multi-turn reasoning process to obtain the final answer. The policy VLM is optimized by the BN-GSPO algorithm, driven by the format reward and answer reward.}
    \label{fig:pipeline}
\end{figure*}

\section{Method}\label{sec:method}
In this section, we investigate how to build an agentic search-reasoning model with fine-grained visual analysis capabilities for complex search-oriented and knowledge-intensive tasks. Our investigation focuses on the two-stage training strategy, data construction, and benchmark evaluation. We first present the formal task formulation in Sec.~\ref{problem_formulation}. We then describe the two-stage training approach in Sec.~\ref{reinforcement_learning}, which consists of cold-start SFT and RL. The data collection pipeline for the SFT and RL stages is detailed in Sec.~\ref{training_data}. Finally, in Sec.~\ref{benchmark_data}, we introduce \datasetname{}, our newly constructed benchmark designed to rigorously evaluate the fine-grained visual analysis and search-reasoning capabilities of VLMs.


\subsection{Problem Formulation}
\label{problem_formulation}
We build upon the MMSearch-R1~\cite{wu2025mmsearch} problem setting for interactive and search-augmented multimodal reasoning in a real web environment, and extend it in two main ways. First, we expand the available toolset beyond text and image search by adding the image crop tool, which allows the agent to zoom into specific regions of an image. Second, we focus on multimodal search tasks that involve high-resolution images, where important information can only be revealed through selective zooming. The left panel of Fig.~\ref{fig:pipeline} presents an overview of this setting. It illustrates how the agent interacts with the environment through reasoning and tool use, and how the task, observation space, and action space are structured within this process.

\noindent \textbf{Task and Objective.}
The task begins with a natural language query $q$ and an initial image $I_0$. At each turn, the agent may invoke a tool or produce a final answer when it deems the information sufficient. The process ends once the final answer is given. If no valid answer is produced within $T$ turns, the result is considered incorrect.

\noindent \textbf{Observation Space.}
At turn $t$, the model observes the full interaction history $\mathcal{T}_t$, which records all text and image elements in order, including the initial prompt, the query, prior reasoning steps, tool calls, and their outputs. Each tool call yields a compact observation $o_t$, in text, image, or both, which is then appended to $\mathcal{T}_t$. For instance, calling the image crop tool adds the resulting cropped image. The framework is agnostic to the specific form of these observations, requiring only that each provides self-contained information for reasoning.

\noindent \textbf{Action Space.}
At each turn, the model first generates a reasoning step, then selects one of four actions:
\begin{enumerate}
    \item Perform a text-based web search using \textit{text search}.
    \item Execute a reverse image search using \textit{image search}.
    \item Zoom into a specific image region using \textit{image crop}.
    \item Produce the final answer.
\end{enumerate}
The output, such as search results or a cropped image, is added to $\mathcal{T}_t$, forming an evolving trajectory. Each turn includes a reasoning step and one valid action. If either is missing, the trajectory is invalid.

\subsection{Reinforcement Learning with Cold-start}
\label{reinforcement_learning}
We follow a two-stage training recipe. In the first stage, we perform cold-start SFT to bootstrap the model's basic ability to learn multi-tool invocations. In the subsequent RL stage, we employ the proposed Batch-normalized GSPO (BN-GSPO) algorithm to further refine the tool invocation and reasoning capabilities.

\noindent \textbf{Cold-start SFT.}
The cold-start stage involves SFT on a small, meticulously curated dataset $\mathcal{D}_{\text{SFT}}$ of multi-turn interaction trajectories. We formulate the cold-start process as follows:
\begin{align}
     \mathcal{L}_{\text{SFT}} = - \sum_{(x_i, y_i) \in \mathcal{D}_{\text{SFT}}} \log \pi_{\theta}(y_i \mid x_i), 
\end{align}
where $\mathcal{D}_{\text{SFT}}$ is the cold-start dataset, $x_i$ denotes the user query, $y_i$ is the target reasoning trajectory, and $\pi_{\theta}$ represents the model parameterized by $\theta$.

\noindent \textbf{BN-GSPO for RL.}
We build on GSPO \cite{zheng2025group} to train agents capable of multimodal search and visual reasoning via external tools. In this setting, each agent output is a sequence that may include text, tool calls, and observations, with supervision provided only at the sequence level through an external reward model. This necessitates sequence-level optimization, for which GSPO is a natural starting point. However, standard GSPO is sensitive to the diverse trajectory structures and reward magnitudes that arise in multimodal and tool-augmented tasks. Different prompts or tool interactions within the same batch can result in varying outcome lengths, reward scales, and difficulty levels, which may bias the learning signal and destabilize training.

To address this issue, we propose BN-GSPO, which applies a two-stage normalization to the advantage estimates. This stabilizes optimization across heterogeneous prompts and preserves consistent learning signals within and across groups. Given a question-answer pair $(x_b, z_b) \sim \mathcal{D}_{RL}$, we sample $G$ responses $\{y_{b,g}\}_{g=1}^G \sim \pi_{\theta_{\text{old}}}(\cdot \mid x_b)$ and define the length-normalized sequence importance ratio:
\begin{equation}
s_{b,g}(\theta) = \left( \frac{\pi_{\theta}(y_{b,g} \mid x_b)}{\pi_{\theta_{\text{old}}}(y_{b,g} \mid x_b)} \right)^{1/|y_{b,g}|} \nonumber = \exp\!\left( \frac{1}{|y_{b,g}|} \sum_{t=1}^{|y_{b,g}|} \log \frac{\pi_{\theta}(y_{b,g,t}\mid x_b, y_{b,g,<t})}{\pi_{\theta_{\text{old}}}(y_{b,g,t}\mid x_b, y_{b,g,<t})} \right)
\end{equation}

Let $r_{b,g}$ denote the scalar sequence-level reward from the external reward model for $(x_b, y_{b,g})$. We first apply GSPO’s group-level standardization to compute the group-normalized rewards:
\begin{equation}
\bar{A}_{b,g} = \frac{r_{b,g} - \mathrm{mean}\left( \{ r_{b,g'} \}_{g'=1}^G \right)}{\mathrm{std}\left( \{ r_{b,g'} \}_{g'=1}^G \right)},
\end{equation}
and then normalize these values across the entire optimizer minibatch:
\begin{equation}
\tilde{A}_{b,g} = \frac{\bar{A}_{b,g} - \mathrm{mean}\!\left( \{ \bar{A}_{b',g'} \}_{b' \in \mathcal{B},\, g' \in \mathcal{G}} \right)}{\mathrm{std}\!\left( \{ \bar{A}_{b',g'} \}_{b' \in \mathcal{B},\, g' \in \mathcal{G}} \right)},
\end{equation}
where $\mathcal{B}$ denotes the current minibatch and $\mathcal{G} = \{1, \dots, G\}$ is the set of group indices. This second step helps correct inconsistent scales and variances across different prompts within the same batch. With the normalized advantages, we apply the following clipped sequence-level objective $J(\theta)$:
\begin{equation}
\begin{split}
&\mathbb{E}_{x_b,\{y_{b,g}\}}\Bigg[ \frac{1}{G} \sum_{g=1}^{G}
\min(s_{b,g}(\theta)\tilde{A}_{b,g}, \mathrm{clip_{\epsilon_{\mathrm{low}}}^{\epsilon_{\mathrm{high}}}}(s_{b,g}(\theta))\tilde{A}_{b,g})
\Bigg]  - \beta\, D_{\mathrm{KL}}(\pi_{\theta}\,\|\,\pi_{\text{ref}}).
\end{split}
\raisetag{12pt}
\end{equation}
This clipped objective stabilizes RL training by preventing excessively large policy updates. Similarly, the KL term helps prevent overfitting by applying a small penalty to deviations from a frozen reference policy $\pi_{\text{ref}}$.

\begin{figure*}[t]
    \centering
    \includegraphics[width=0.95\linewidth]{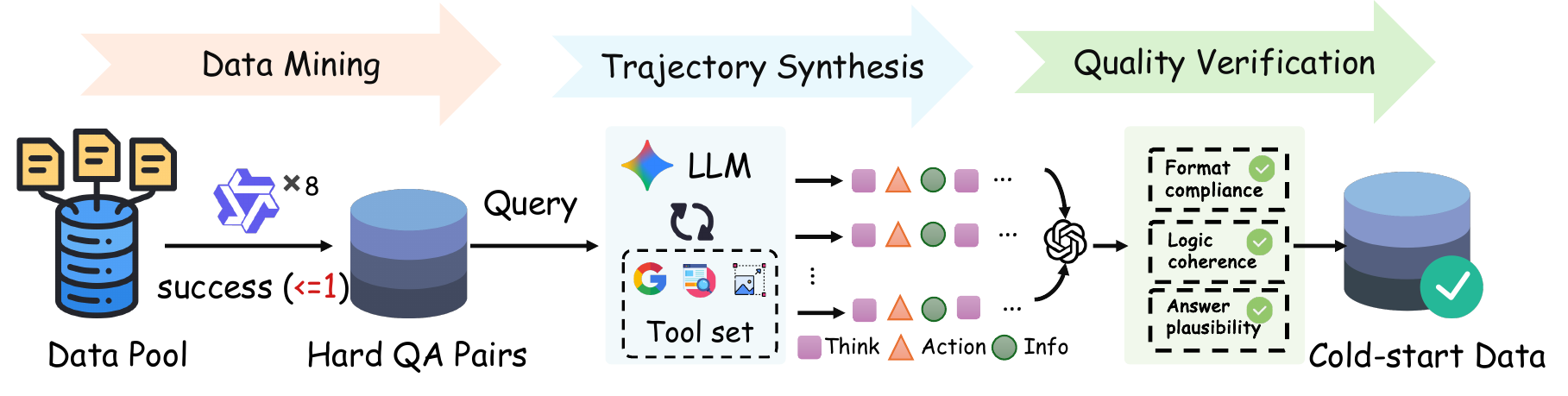}
    \caption{Cold-start data generation pipeline. It consists of data mining, trajectory synthesis and quality verification.}  \label{fig:cold_start_pipeline}
    \vspace{-10pt}
\end{figure*}

\noindent \textbf{Reward Modeling.}
To optimize the learning process in BN-GSPO, we formulate the sequence-level reward as a combination of final answer accuracy and structural format compliance.
Specifically, for a trajectory $\tau$, the total sequence-level reward $R(\tau)$ is given by $R(\tau) = R_{acc}(\tau) + R_{format}(\tau).$
The accuracy reward $R_{acc}(\tau)$ measures the semantic agreement between the predicted answer and the ground truth, which is evaluated using an LLM-as-a-judge.
The format reward $R_{format}(\tau)$ guarantees strict compliance with the interaction protocol. Under this protocol, each non-final turn must consist of a reasoning trace and a single tool call. The final turn, in contrast, must contain the reasoning trace and the answer.
Additionally, the protocol requires all content to be enclosed within special tags and all tool calls to conform to a specified JSON schema. Full details are provided in the Appendix.

    

\subsection{Training Dataset Collection}
\label{training_data}
For cold-start SFT, we design a structured three-phase data pipeline, as illustrated in Fig.~\ref{fig:cold_start_pipeline}. The pipeline begins with data filtering, where we merge the FVQA train set~\cite{wu2025mmsearch}, the Pixel-Reasoner warm-start corpus~\cite{su2025pixel}, and curated expert-annotated multimodal QA pairs to construct the raw data pool. We then filter this pool using 8 rollouts from Qwen2.5-VL-7B-Instruct, marking a sample as \textit{hard} if the model answers it correctly one time or fewer. For these hard QA samples, we prompt Gemini-2.5-Flash to synthesize complete solution trajectories through tool invocations. Finally, GPT-4o is used to verify format compliance, logical coherence, and answer plausibility. Only the validated results are retained, yielding \textasciitilde{}3{,}000 high-quality samples. For RL, we use FVQA-train \cite{wu2025mmsearch} together with VisualProbe-train and DeepEyes-4K-train \cite{lai2025mini}, which provides a robust mix of factual question answering and high-resolution visual analysis tasks.

\subsection{\datasetname{} Benchmark Construction}
\label{benchmark_data}


\begin{figure}
    \centering
    \includegraphics[width=0.65\linewidth]{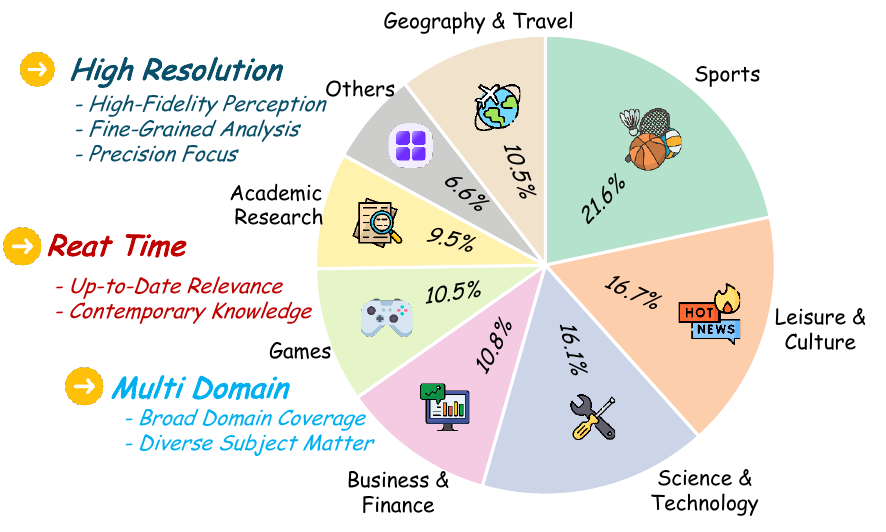}
    \caption{Statistics of our proposed HR-MMSearch benchmark. HR-MMSearch is characterized by the high-resolution images and knowledge-intensive question, covering areas such as Sports, Leisure\&Culture,  Science\&Technology, Business\&Finance, Games, and Academic Research.}  
    \label{fig:hr_mmsearch}
\end{figure}

Existing benchmarks, such as FVQA \cite{wu2025mmsearch} or MMSearch \cite{jiang2024mmsearch}, typically utilize standard HD or lower-resolution images and focus on holistic scene understanding, which leaves a critical gap in evaluating an agent's detailed visual understanding capabilities. Thus, we introduce \datasetname{} as shown in Fig.~\ref{fig:hr_mmsearch}, a benchmark for the fine-grained perception and search-reasoning capabilities of VLM agents. This dataset consists of 305 4K-resolution images curated from 8 diverse, high-impact domains, covering areas such as Sports, Leisure \& Culture, and Science \& Technology. To prevent data leakage from the VLMs' pre-trained knowledge, all images are sourced exclusively from recent 2025 events. For each image, we manually craft knowledge-intensive, search-oriented questions that focus on a key visual subject, such as a small or inconspicuous object or text occupying less than 5\% of the image area.

\begin{table*}[t]
\caption{Performance on search-oriented benchmarks under Direct Answer and Agentic Model workflows. 
}
\label{tab:agentic_search}
\setlength{\tabcolsep}{3pt}
\centering
\resizebox{\textwidth}{!}{%
\begin{tabular}{c|c|c|ccccccc}
\toprule
\textbf{Type} & \textbf{Model} & \textbf{Average} & \textbf{MMSearch} & \textbf{\datasetname{}} & \textbf{FVQA-test} & \textbf{InfoSeek} & \textbf{SimpleVQA} & \textbf{LiveVQA} & \textbf{MAT-Search} \\
\midrule
\multicolumn{10}{c}{\textit{\textbf{Direct Answer}}} \\ \midrule
\multirow{3}{*}{\shortstack{Open-source}} 
 & Qwen2.5-VL-7B-Instruct \cite{bai2025qwen2}    & 27.70 & 7.60  & 0.58  & 26.28 & 31.95 & 47.88 & 19.63 & 60.00 \\
 & Qwen3-VL-8B-Instruct \cite{Qwen3-VL}          & 29.24 & 11.70 & 12.13 & 24.22 & 23.15 & 42.94 & 23.18 & 67.33 \\
 & Qwen2.5-VL-32B-Instruct \cite{bai2025qwen2}  & 32.01 & 11.70 & 3.93  & 30.50 & 36.65 & 48.57 & 21.40 & 71.33 \\
 & Qwen3-VL-32B-Instruct \cite{Qwen3-VL}  & 35.22 & 16.96 & 19.02  & 32.17 & 28.95 & 45.90 & 31.59 & 72.67 \\
 \midrule
\multirow{5}{*}{Proprietary} 
 & GPT-4o-mini \cite{hurst2024gpt}              & 33.08 & 15.79 & 1.31  & 36.83 & 35.95 & 44.42 & 24.63 & 72.66 \\
 & Gemini-2.5-Flash \cite{comanici2025gemini}   & 40.87 & 21.64 & 7.54  & 43.78 & 44.10 & 55.48 & 31.57 & 82.00 \\
 & GPT-4o \cite{hurst2024gpt}                    & 42.38 & 23.39 & 13.11 & 48.00 & 52.90 & 51.73 & 28.18 & 79.33 \\
 & GPT-5 \cite{openai2025gpt5}                  & 50.24 & 35.09 & 22.62 & 54.39 & 54.15 & 61.70 & 44.39 & 79.33 \\
  & GPT-5.2 \cite{openai2025gpt5}                  & 50.92 & 43.27 & 24.92 & 50.94 & 50.40 & 59.92 & 47.00 & 80.00 \\
 & Gemini-3-Flash \cite{gemini3flash}            & 53.68 & 57.31 & 21.97 & 56.50 & 54.85 & 63.57 & 38.90 & 82.67 \\
& Gemini-3-Pro \cite{gemini3pro}            & 55.87 & 62.57 & 26.89 & 59.22 & 56.30  & 64.07 & 40.06 & 82.00 \\
\midrule
\multicolumn{10}{c}{\textit{\textbf{Agentic Model (zero-shot)}}} \\ \midrule
\multirow{3}{*}{\shortstack{Open-source}} 
 & Qwen2.5-VL-7B-Instruct \cite{bai2025qwen2}    & 35.50 & 32.16 & 19.34 & 36.00 & 28.80 & 42.35 & 22.52 & 67.33 \\
& Qwen3-VL-8B-Instruct \cite{Qwen3-VL}          & 51.52 & 47.37 & 27.87 & 53.61 & 46.15 & 62.29 & 39.37 & 84.00 \\ 
 & Qwen2.5-VL-32B-Instruct \cite{bai2025qwen2}  & 53.45 & 49.71 & 33.44 & 52.22 & 50.10 & 65.15 & 42.17 & 81.33 \\
 & Qwen3-VL-32B-Instruct \cite{Qwen3-VL}  & 53.82 & 49.12 & 34.43 & 54.28 & 49.85 & 64.17 & 42.87 & 82.00 \\ \midrule
\multirow{5}{*}{Proprietary} 
 & GPT-4o-mini \cite{hurst2024gpt}              & 45.65 & 38.60 & 26.23 & 50.00 & 42.35 & 50.84 & 31.54 & 80.00 \\
  & GPT-4o \cite{hurst2024gpt}                    & 55.09 & 49.12 & 30.16 & 66.34 & 59.55 & 63.67 & 40.09 & 76.67 \\
 & Gemini-2.5-Flash \cite{comanici2025gemini}   & 58.05 & 59.06 & 40.00 & 61.72 & 53.70 & 68.81 & 47.75 & 75.33 \\
 & GPT-5 \cite{openai2025gpt5}                  & 60.12 & 52.63 & 38.36 & 62.61 & 55.95 & 70.58 & 56.02 & 84.67 \\
 & Gemini-3-Flash \cite{gemini3flash}            & 61.26 & 62.57 & 41.64 & 64.89 & 61.10 & 67.92 & 48.06 & 82.67 \\
  & GPT-5.2 \cite{openai2025gpt5}                 & 67.64 & 66.08 & 48.20 & 68.78 & 65.55 & 78.18 & 65.99 & 80.67\\
 & Gemini-3-Pro \cite{gemini3pro}            & 69.06 & 74.27 & 48.52 & 72.61 & 66.45 & 75.91 & 59.69 & 86.00 \\
\midrule
\multicolumn{10}{c}{\textit{\textbf{Agentic Model}}} \\ \midrule
\multirow{6}{*}{\shortstack{Open-source}} 
 & Visual-ARFT \cite{liu2025visual}             & 40.13 & 34.50 & 24.92 & 41.72 & 37.95 & 42.45 & 25.40 & 74.00 \\
 & DeepMMSearch-R1 \cite{narayan2025deepmmsearch} & --    & --    & --    & --    & 47.51 & 55.87 & --    & -- \\
 & MMSearch-R1 \cite{wu2025mmsearch}             & 52.49 & 53.80 & 20.33 & 58.40 & 55.10 & 57.40 & 48.40 & 74.00 \\
 & DeepEyesV2 \cite{hong2025deepeyesv2}          & --    & 63.70 & --    & 60.60 & 51.10 & 59.40 & --    & -- \\
 & \cellcolor{green!10}\textbf{\methodname{}-8B}         & \cellcolor{green!10}64.20 & \cellcolor{green!10}67.84 & \cellcolor{green!10}41.64 & \cellcolor{green!10}67.11 & \cellcolor{green!10}61.70 & \cellcolor{green!10}70.19 & \cellcolor{green!10}56.22 & \cellcolor{green!10}84.67 \\
  & \cellcolor{green!10}\textbf{\methodname{}-32B}         & \cellcolor{green!10}\textbf{69.74} & \cellcolor{green!10}\textbf{74.27} & \cellcolor{green!10}\textbf{54.43} & \cellcolor{green!10}\textbf{72.61} & \cellcolor{green!10}\textbf{65.25} & \cellcolor{green!10}\textbf{74.14} & \cellcolor{green!10}\textbf{60.83} & \cellcolor{green!10}\textbf{86.67} \\
\bottomrule
\end{tabular}%
}
\end{table*}

\section{Experiments}
\label{sec:experiments}
\subsection{Implementation Details}
\noindent \textbf{Model and Training.} \methodname{}-7B, which is based on Qwen2.5-VL-7B-Instruct \cite{bai2025qwen2}, is trained using the two-stage pipeline.
This process involves SFT followed by RL, which we implement using the LLaMA-Factory \cite{zheng2024llamafactory} and veRL \cite{sheng2025hybridflow} frameworks, respectively. 
During the SFT stage, we fine-tune only the language model while keeping the vision encoder and multi-modal projector frozen. The training uses a learning rate of $1 \times 10^{-5}$ for 3 epochs. The SFT stage bootstraps the model's ability to follow the interaction protocol and utilize the multi-toolset. The subsequent RL stage then teaches the model to use these tools more effectively and efficiently. For RL stage, we use a global batch size of 128, a learning rate of $1\times10^{-6}$, and a KL coefficient $\beta$ of $1\times10^{-4}$. Additionally, to encourage better exploration, we follow DAPO \cite{yu2025dapo} by using the Clip-Higher strategy with $\epsilon_{\text{low}} = 0.2$ and $\epsilon_{\text{high}} = 0.28$. During this RL phase, a single training trajectory allows the agent to interact for a maximum of $T=10$ turns. This means the agent can iteratively use tools and refine its plan for up to 10 steps before producing a final answer for that trajectory. In each turn, the agent can generate up to $8,192$ tokens, with a cumulative limit of $32,768$ tokens for the entire trajectory. 
Compared with \methodname{}-7B, \methodname{}-8B and \methodname{}-32B are developed based on Qwen3-VL-8B-Instruct and Qwen3-VL-32B-Instruct \cite{Qwen3-VL}, respectively, and only the RL stage is employed, leveraging the strong tool-usage capability of the base model.
Detailed training hyperparameters are available in the Appendix.

\noindent \textbf{Training Rewards.} 
To guide the RL process, we employ a two-component reward that jointly promotes answer correctness and format compliance. Each component is a binary score determined by GPT-4o acting as the LLM-as-a-Judge. 
Specifically, the accuracy reward $R_{acc}(\tau)=1.0$ and format reward $R_{format}(\tau)=0.5$ are conferred when their respective criteria are fully satisfied, and are 0.0 otherwise.

\noindent \textbf{Multimodal Tools.} \methodname{} can invoke following three multimodal tools using JSON-style arguments:

\begin{itemize}
    \item \textit{text search}: This tool enables the agent to query the web for textual information. It accepts a single text query as its argument and is powered by the Serper Search API. Following MMSearch-R1 \cite{wu2025mmsearch}, to avoid overwhelming the context length of the model, the top five results are first summarized by Qwen3-32B \cite{yang2025qwen3} before being returned to the agent.
    \item \textit{image search}: This tool performs a reverse image search to retrieve images that are visually similar or contextually related to a given input. It takes an image index as its argument and is powered by the Serper Image Search API. To minimize financial cost and reduce latency during RL training, the top five image search titles and thumbnails for all prompts in the training dataset are pre-fetched and cached in advance.
    \item \textit{image crop}: This tool allows the agent to crop a previously seen image, allowing focused fine-grained analysis of a specific region of interest. It requires two arguments: normalized coordinates of bounding box $[0.0, 1.0]$ and an index referencing the target image.
\end{itemize}

\noindent \textbf{Evaluation Benchmarks.} To evaluate these capabilities, we test the model on two corresponding categories of benchmarks. For agentic search, we use benchmarks including FVQA-test~\cite{wu2025mmsearch}, InfoSeek~\cite{chen2023can}, MMSearch~\cite{jiang2024mmsearch}, SimpleVQA~\cite{cheng2025simplevqa}, LiveVQA~\cite{fu2025livevqa}, MAT-Search~\cite{liu2025visual}, and our newly constructed HR-MMSearch benchmark. For visual reasoning, we test fine-grained understanding capability using V* Bench~\cite{wu2024v}, HR-Bench~\cite{wang2025divide} and MME Realworld~\cite{zhang2024mme}. Full details on all these datasets are provided in the Appendix.

\noindent \textbf{Baselines.} We evaluate our model against several strong baselines. These include proprietary models, such as GPT-5 \cite{openai2025gpt5}, GPT-5.2 \cite{openai2025gpt5_2}, Gemini-3-Flash \cite{gemini3flash} and Gemini-3-Pro \cite{gemini3pro}, as well as open-source agentic models like MMSearch-R1 \cite{wu2025mmsearch}, DeepMMSearch-R1 \cite{narayan2025deepmmsearch} and DeepEyesV2 \cite{hong2025deepeyesv2}. All models are tested under the following two workflow settings: 
\begin{itemize} 
\item \textit{Direct Answer}: The model produces an answer directly without using external tools. 
\item \textit{Agentic Model}: The model is provided with all three tools and autonomously decides how to use these tools in the rollout reasoning process. 
\end{itemize} 
Details for these workflows are provided in the Appendix.

\noindent \textbf{Evaluation Metrics.}  For agentic search tasks, the primary metric is Pass@1, assessed by a GPT-4o judge that scores the model's final answer against ground-truth results.
For visual understanding benchmarks, we report Avg@8 Exact Match on V* Bench and HR-Bench, and Pass@1 Exact Match on the large-scale MME-RealWorld benchmark. The evaluation prompt is provided in the Appendix.

\begin{table}[t]
    \centering
    \caption{Performance on visual understanding benchmarks.}
    \label{tab:crop_bench_performance}
    \begin{tabular}{l | cccc|c}
        \toprule
        \textbf{Model} & \textbf{V* Bench} &  \textbf{HR-Bench 4K} &  \textbf{HR-Bench 8K} & \textbf{MME RealWorld} & \textbf{Avg.} 
        \\
        \midrule
        \multicolumn{6}{c}{\textit{\textbf{Direct Answer}}} \\
        \midrule
        Gemini-2.5-Pro~\cite{comanici2025gemini} & 83.8 & 87.3 & 85.4 & - & - \\  
        GPT-4o~\citep{hurst2024gpt}  & 67.5 & 65.0 & 59.6 & 62.8 & 63.7 \\
        LLaVA-onevison~\citep{li2024llava} & 75.4 & 63.0 & 59.8 & 57.4 & 63.9 \\
        Qwen2.5-VL-7B-Instruct~\citep{bai2025qwen2} & 75.3 & 65.5 & 62.1 & 56.8 & 64.9 \\
        Qwen2.5-VL-32B-Instruct~\citep{bai2025qwen2} & 80.6 & 69.3 & 63.6 & 59.1 & 68.2 \\
        Qwen3-VL-8B-Instruct~\citep{Qwen3-VL} & 86.4 & 78.9 & 74.6 & 61.9 & 75.5 \\
        \midrule
        \multicolumn{6}{c}{\textit{\textbf{Agentic Model}}} \\
        \midrule
        SEAL~\citep{wu2024v} & 74.8 & - & - & - & - \\
        Qwen3-VL-32B-Instruct~\citep{Qwen3-VL} & 91.1 & 84.6 & 81.6 & - & - \\
        Qwen3-VL-235B-A22B-Instruct~\citep{Qwen3-VL} & 93.7 & 85.4 & 82.4 & - & - \\
        Monet~\citep{wang2025monet} & 83.3 & 71.0 & 68.0 & - & - \\
        Pixel-Reasoner~\citep{su2025pixel}  & 84.3 & 72.6 & 66.1 & 64.4 & 71.9 \\
        DeepEyes~\citep{zheng2025deepeyes}  & 83.3 & 73.2 & 69.5 & 64.1 & 72.5 \\
        Thyme~\cite{zhang2025thyme} & 82.2 & 77.0 & 72.0 & 64.8 & 74.0 \\
        DeepEyesV2~\cite{hong2025deepeyesv2}  & 81.8 & 77.9 & 73.8 & 64.9 & 74.6 \\
        Mini o3~\citep{lai2025mini}  & 88.2 & 77.5 & 73.3 & 65.5 & 76.1 \\
        Skywork-R1V4~\citep{zhang2025skywork}  & 88.0 & 82.8 & 79.8 & 71.4 & 80.5 \\

        \rowcolor{green!10}
        \textbf{\methodname{}-8B} & 92.2 & 83.1 & 78.4 & 67.9 & 80.4 \\
        \rowcolor{green!10}
        \textbf{\methodname{}-32B} & \textbf{94.2} & \textbf{90.2} & \textbf{86.6} & \textbf{72.7} & \textbf{85.9} \\
        \bottomrule
    \end{tabular}
\end{table}

\subsection{Main Results}

\begin{figure}[t]
    \centering
    \includegraphics[width=\linewidth]{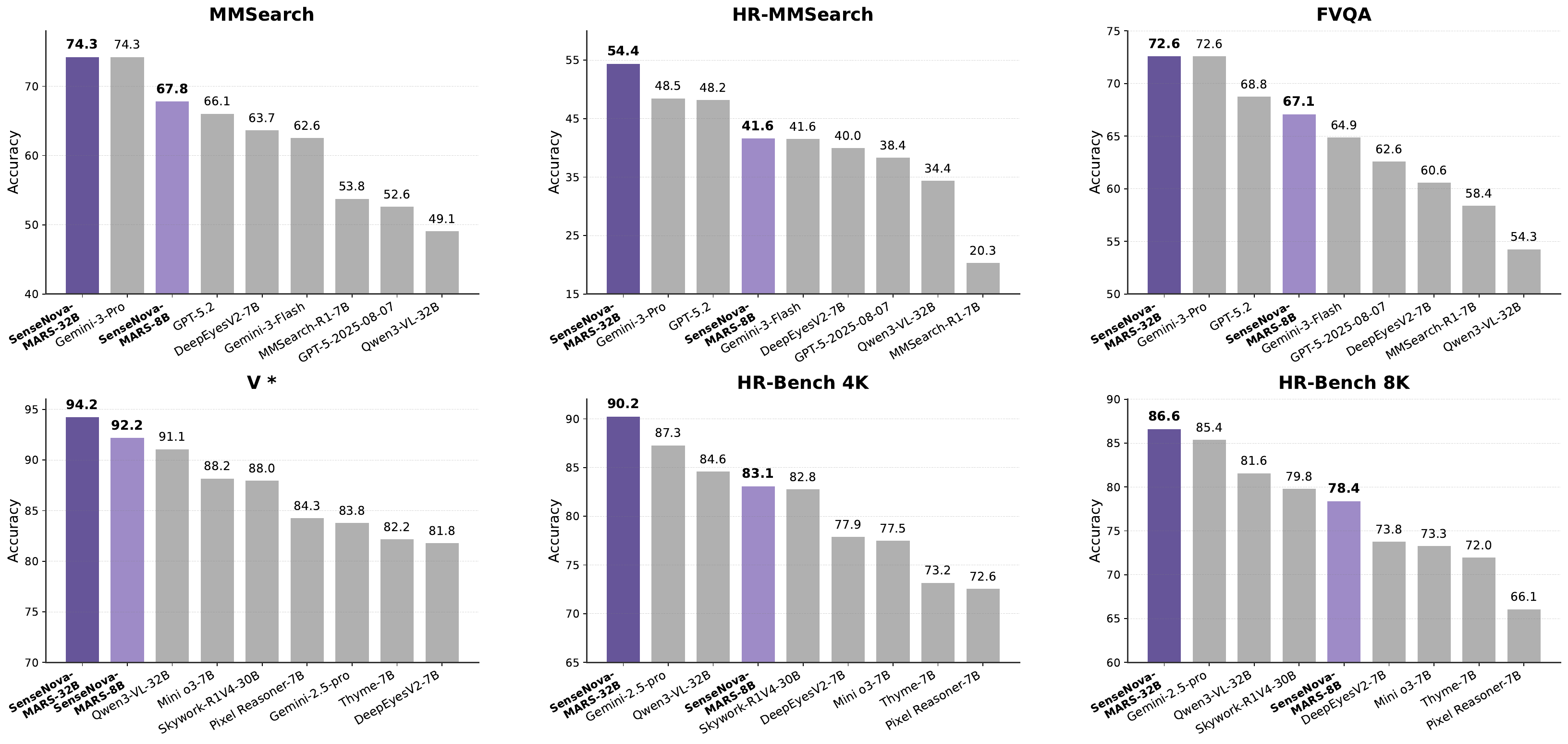}
    \caption{Overall performance of SenseNova-MARS-32B and  SenseNova-MARS-8B compares to other models.}  
    \label{fig:fig_compare_32b_8b}
\end{figure}

\noindent \textbf{Search-oriented Evaluation.} As shown in Tab.~\ref{tab:agentic_search} and Fig.~\ref{fig:fig_compare_32b_8b}, \methodname{}-32B establishes a new SOTA among open-source agentic models and closed-source proprietary models.
\methodname{}-32B achieves an average score of 69.74 across seven benchmarks,  surpassing leading proprietary models such as  Gemini-3-Pro \cite{gemini3pro} and GPT-5.2 \cite{openai2025gpt5_2}.
Notably, on the HR-MMSearch benchmark, \methodname{}-32B scores 54.43, outperforming Gemini-3-Pro and GPT-5.2 by margins of 5.91 and 6.23 points, respectively.
In the small-scale category, \methodname{}-8B also demonstrates strong performance, attaining an average gain of 12.68 points over Qwen3-VL-8B.
Additionally, \methodname{} outperforms search-oriented agentic models such as MMSearch-R1 \cite{wu2025mmsearch}, DeepMMSearch-R1 \cite{narayan2025deepmmsearch}, and DeepEyesV2 \cite{hong2025deepeyesv2}, surpassing MMSearch-R1 by an average of 11.71 points.
Compared with proprietary closed-source models, \methodname{}-8B surpasses GPT-5 \cite{openai2025gpt5}, Gemini-2.5-Flash \cite{comanici2025gemini}, and Gemini-3-Flash \cite{gemini3flash} by a significant margin, outperforming Gemini-3-Flash by an average of 2.94 points. 
These results highlight the efficacy of our unified agentic search and visual reasoning RL framework.

\noindent \textbf{Fine-grained Visual Understanding.} As shown in Tab.~\ref{tab:crop_bench_performance} and Fig.~\ref{fig:fig_compare_32b_8b}, \methodname{} demonstrates superior fine-grained perception capabilities on high-resolution benchmarks, validating the effectiveness of our approach in detailed, pixel-space analysis.
\methodname{}-32B scores 94.2 on  V* Bench \cite{wu2024v} and 90.2 on HR-Bench 4k \cite{wang2025divide}, surpassing top-tier models such as Qwen3-VL-235B-A22B-Instruct \cite{Qwen3-VL} and Gemini-2.5-Pro \cite{comanici2025gemini}.
\methodname{}-8B achieves leading scores of 92.2 on V* Bench \cite{wu2024v}, 83.1 on HR-Bench 4k \cite{wang2025divide}, 78.4 on HR-Bench-8k \cite{wang2025divide}, and 67.9 on MME-RealWorld \cite{zhang2024mme}, outperforming all existing tool-based models such as Pixel Reasoner~\cite{su2025pixel}, DeepEyes \cite{zheng2025deepeyes}, and Mini o3 \cite{lai2025mini}.
Compared with Qwen3-VL-8B, \methodname{}-8B attains an average performance gain of 4.9 points, demonstrating the effectiveness of our proposed RL algorithm.

\subsection{Ablation Study}


\noindent \textbf{Effectiveness of Proposed BN-GSPO.} We evaluate the effectiveness of our proposed BN-GSPO based on \methodname{}-7B by comparing it against the commonly used RL algorithms GRPO and GSPO in Tab.~\ref{comparison_rl_algorithm}. To isolate the influence of SFT, this comparison uses a pure RL setup where all models are initialized from Qwen2.5-VL-7B-Instruct~\cite{bai2025qwen2} without a cold-start. As shown, BN-GSPO achieves the best overall performance across all benchmarks and demonstrates stable, balanced improvements. In contrast, while GRPO and GSPO improve on some tasks, they fail to perform well across both search and perception. This indicates that batch normalization effectively mitigates reward scale variance, leading to more consistent and robust multi-tool RL training.


\begin{figure}[t]
    \centering
    \includegraphics[width=0.8\linewidth]{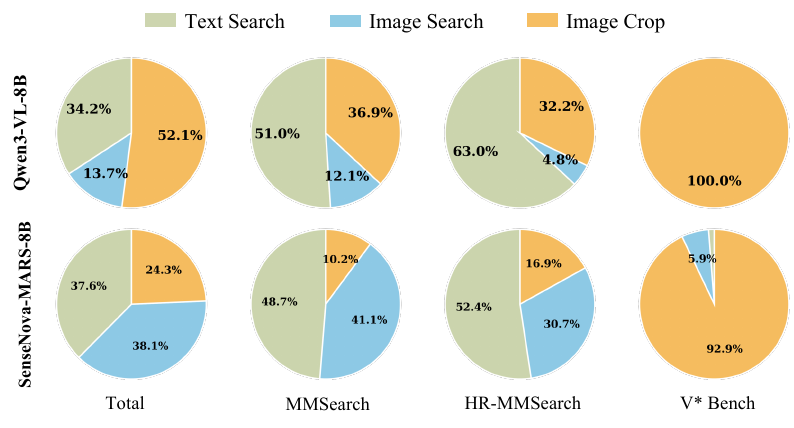}
    \caption{Distribution of tool calls across different benchmarks for Qwen3-VL-8B and \methodname{}-8B.}  
    \label{fig: tool_use_qwen3}
\end{figure}

\begin{table}[H]
    \centering
    \caption{Effectiveness of Proposed BN-GSPO for \methodname{}-7B.}
    \label{comparison_rl_algorithm}
    \begin{tabular}{l|ccc}
        \toprule
        \textbf{Method} & \textbf{MMSearch} & \textbf{V* Bench} & \textbf{HR-Bench 4K} \\
        \midrule
        GRPO \cite{shao2024deepseekmath}  & 50.88 & 67.54 & 61.38 \\
        GSPO \cite{zheng2025group}        & 53.80 & 53.93 & 44.50 \\
        \rowcolor{green!10}
        \textbf{BN-GSPO} & \textbf{56.72} & \textbf{79.05} & \textbf{69.12} \\
        \bottomrule
    \end{tabular}
\end{table}

\begin{table}[H]
    \centering
    \caption{Impact of different data distribution on RL performance for \methodname{}-7B.}
    \label{tab:dataset_usage}
    \begin{tabular}{cc|ccc}
        \toprule
        \textbf{Search} & \textbf{Perception} & \textbf{MMSearch} & \textbf{HR-MMSearch} & \textbf{V* Bench} \\
        \midrule
        \multicolumn{2}{c|}{\textbf{\methodname{}-7B-SFT}}   & 53.80 & 29.80 & 82.20 \\
        \midrule
        \checkmark            &             & 54.97 & 36.80 & 82.72 \\
                              & \checkmark       & 54.09 & 33.11 & \textbf{85.24} \\
        \rowcolor{green!10}
        \checkmark        & \checkmark       & \textbf{59.06} & \textbf{38.52} & 83.84 \\
        \bottomrule
    \end{tabular}
\end{table}

\noindent \textbf{Impact of different data distributions on RL performance.} We evaluate the impact of RL data distributions for \methodname{}-7B in Tab.~\ref{tab:dataset_usage}. Training the model only on specialized data, such as fine-grained perception datasets, makes the agent overspecialize. This boosts V* Bench performance to 85.24 but causes a major drop on search-oriented tasks compared to the SFT baseline. Using our full hybrid dataset, which includes both search and perception tasks, gives the best results by a wide margin on search-oriented metrics. These findings show that hybrid training data is essential. It provides the signals the agent needs to learn a unified, multi-tool policy that avoids overspecialization and balances external knowledge retrieval with fine-grained visual analysis.

\noindent \textbf{Analysis of tool use behavior.} Figure~\ref{fig: tool_use_qwen3} illustrates the adaptive tool-use behavior of \methodname{}-8B.
On the knowledge-intensive MMSearch benchmark, \methodname{} primarily relies on image and text search tools to acquire external information, with minimal dependence on cropping-based perception.
In contrast, on the more challenging \datasetname{}, which requires both high-resolution perception and complex reasoning, \methodname{} demonstrates a more balanced tool usage, indicating effective integration of localized visual cues and external knowledge.
Overall, compared to Qwen3-VL-8B, \methodname{}-8B shows stronger adaptability across diverse tasks by dynamically selecting the most effective tools for each task setting.

\section{Conclusion}\label{sec:conclusion}
In this work, we propose \methodname{}, a novel multimodal agentic reasoning and search framework that can actively employ the image search, text search and image crop tools within the multi-turn reasoning process for search-driven and fine-grained visual tasks. 
To empower this capability, we introduce the BN-GSPO algorithm to enhance the reasoning robustness and tool-use proficiency.
Furthermore, we construct the HR-MMSearch Benchmark with high-resolution images and knowledge-intensive questions to evaluate the performance of VLMs. 
Extensive experiments demonstrate that \methodname{} achieves superior performance across diverse benchmarks, showcasing efficient tool invocation and robust reasoning capabilities.

\clearpage

\bibliographystyle{plainnat}
\bibliography{main}

\clearpage


\beginappendix

\section{Additional Details on Training Data}\label{sec:traing data}
In this section, we provide more details on the training data used for both the cold-start supervised fine-tuning (SFT) phase and the reinforcement learning (RL) phase.

\subsection{Cold-Start SFT Data}
The Cold-start SFT dataset is constructed to equip the model with foundational capabilities in tool usage and reasoning. As visualized in the main text, we apply a rigorous filtering and synthesis pipeline to three primary data sources:

\begin{itemize}
    \item \textbf{FVQA~\cite{wu2025mmsearch}:} Starting from the original training set of 4,849 samples, we identify "hard" samples where the base model frequently failed. After trajectory synthesis and validation, we retain \textbf{1,115} high-quality trajectories for SFT.
    \item \textbf{Pixel-Reasoner Corpus~\cite{su2025pixel}:} We leverage the warm-start corpus containing 7.85k samples. Through our filtering process, we select approximately \textbf{2,000} samples that best demonstrate pixel-level reasoning capabilities.
    \item \textbf{Curated Expert Data:} To specifically enhance the model's proficiency in complex visual scene with multi-step tool invocations, we manually construct the training subset comprising \textbf{200} intricate reasoning trajectories.
\end{itemize}
This process yields a total of approximately 3,315 high-quality samples for the cold-start SFT phase.

\subsection{Reinforcement Learning Data}
For the RL phase, we utilize a larger and more diverse dataset to generalize the model's reasoning and tool-use policies. The RL training set comprises:
\begin{itemize}
    \item \textbf{FVQA (Remaining):} The subset of the FVQA training set not used for SFT, consisting of \textbf{3,695} samples.
    \item \textbf{DeepEyes-4K~\cite{lai2025mini}:} We utilize \textbf{4,000} samples from the DeepEyes-4K training set to reinforce high-resolution visual analysis.
    \item \textbf{Visual-Probe~\cite{lai2025mini}:} We include the complete training set of \textbf{5,729} samples to support broad visual reasoning tasks.
\end{itemize}

\section{Additional Details on Evaluation Data}\label{sec:evaluation_data}
In this section, we provide more details on the evaluation data used for agentic search and visual reasoning task.

\subsection{Agentic Search}
For agentic search, we mainly rely on our proposed HR-MMSearch, along with MMSearch \cite{jiang2024mmsearch}, FVQA-test \cite{wu2025mmsearch}, Infoseek \cite{chen2023can}, SimpleVQA \cite{cheng2025simplevqa}, LiveVQA \cite{fu2025livevqa} and MAT-Search \cite{liu2025visual}:

\begin{figure*}[t]
    \centering
    \includegraphics[width=\linewidth]{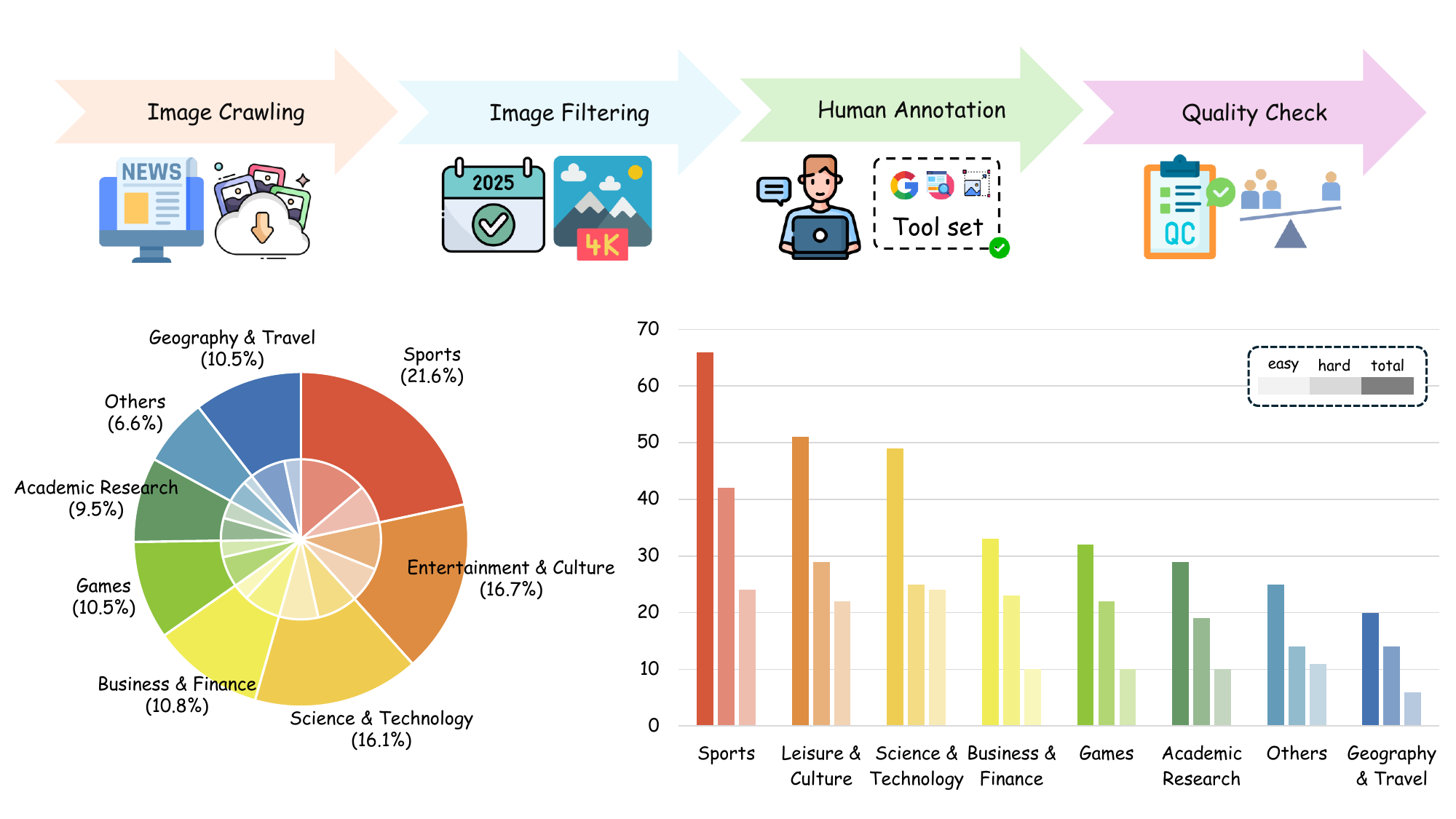}
    \caption{Overview of the proposed \datasetname{} dataset. This figure details the methodology used to construct the dataset and presents the resulting distribution of categories and difficulties. The colorbar uses a gradation of dark, medium, and light to denote the number of total, hard, and easy samples, respectively. } 
    \label{fig:benchmark_construction}
\end{figure*}

\begin{itemize}
\item \textbf{HR-MMSearch.}
Existing benchmarks, such as FVQA-test \cite{wu2025mmsearch} or MMSearch \cite{jiang2024mmsearch}, typically rely on standard HD or lower-resolution images to test holistic scene understanding. However, this approach leaves a critical gap in evaluating an agent’s ability to understand fine visual details. To fill this gap, we introduce \datasetname{}, a benchmark designed to assess the fine-grained perception and search-reasoning capabilities of vision-language model (VLM) agents. As shown in Fig.~\ref{fig:benchmark_construction}, we construct the dataset through a four-phase pipeline: large-scale image crawling, filtering, human annotation, and rigorous quality checking. We begin by crawling candidate images exclusively from three reputable international news outlets—Reuters, the Associated Press (AP), and CNBC. By focusing on timely news photographs, we ensure that the images depict recent events unlikely to appear in existing VLM pre-training data. This design choice reduces the chance of models relying on memorized knowledge and instead encourages genuine use of external tools. After collection, we apply strict filtering to retain only 4K-resolution images from 2025, minimizing pre-training leakage risks while providing rich, fine-grained visual detail. During the human annotation phase, three annotators (all holding bachelor’s degrees) independently assign each image to one of eight high-impact domains: Sports, Entertainment \& Culture, Science \& Technology, Business \& Finance, Games, Academic Research, Geography \& Travel, and Others. They then manually craft knowledge-intensive questions that target key visual subjects, especially small objects or text regions that occupy less than 5\% of the image. Each question is designed so that solving it requires at least one of three multimodal tools: image search, text search, or image crop. Afterwards, three additional experts, each holding at least a master’s degree, carefully cross-verify the resulting \textbf{305} image–question pairs to confirm the labels, assess quality, and ensure answer correctness. To define difficulty levels, we adopt a pass@8 evaluation protocol in which the agent generates eight independent rollouts for each question using the available tools. A question is considered solved if at least one rollout yields the correct answer. Using this agentic setup, we run Qwen2.5-VL-7B-Instruct as a representative agent to approximate question difficulty. Based on its performance, we categorize the 188 questions for which the model fails all eight rollouts as \textit{Hard}. These failures typically arise on questions that require more complex reasoning or interaction, often involving three or more tool calls; notably, 17 of these questions require coordinated use of all three tools. The remaining 117 questions are classified as \textit{Easy}, as the model produces at least one correct rollout. These easier questions generally require only one or two tool calls. As shown in the bottom-right panel of Fig.~\ref{fig:benchmark_construction}, the difficulty distribution is roughly consistent across all eight categories, with around 60\% \textit{Hard} samples and 40\% \textit{Easy} samples in each. Overall, \datasetname{} offers a challenging and diverse benchmark for evaluating the capabilities of tool-augmented VLM agents in agentic search and fine-grained visual reasoning.

\item \textbf{MMSearch.} We use MMSearch \cite{jiang2024mmsearch} to test whether models can retrieve up-to-date information or reason about obscure facts. The full dataset contains 300 manually collected examples across 14 subdomains, split into News and Knowledge sections. The News section covers events starting from August 2024 to avoid overlap with training data, while the Knowledge section focuses on rare facts that often challenge advanced models. Similar to MMSearch-R1 \cite{wu2025mmsearch}, we only use the \textbf{171} questions that include images and exclude text-only queries to focus on real-world information seeking with visual grounding.

\item \textbf{FVQA-test.} We use the FVQA-test set \cite{wu2025mmsearch} to ensure our evaluation spans both visual and textual domains. This benchmark includes \textbf{1,800} high-quality examples from three sources. The first 600 come from FVQA-auto-vc and are verified for accuracy and separated from training data. Another 600 are taken from the InfoSeek Human Split, with manually corrected answers to fix missing public labels. The final 600 were created by human annotators specifically for this benchmark to expand its coverage.

\item \textbf{InfoSeek.} We evaluate real-world knowledge retrieval using the InfoSeek benchmark \cite{chen2023can}. Its creators generated questions by converting Wikidata triples into natural-language questions using human-designed templates. These templates were developed for 300 relations and include placeholders for units and entity types to improve clarity. They removed unanswerable questions and balanced the dataset across entities to ensure quality. From the test split, we sample \textbf{2,000} instances to capture a diverse set of factual queries.

\item \textbf{SimpleVQA.} SimpleVQA \cite{cheng2025simplevqa} focuses on factual accuracy and real-world usefulness. It combines two types of examples: image--question pairs from post-2023 VQA datasets and new pairs produced by experts using internet search results. All examples pass strict filters for difficulty and quality to ensure they test objective information. From the full benchmark, we use the \textbf{1,013} English examples to evaluate factual reasoning without language-related noise.

\item \textbf{LiveVQA.} To measure performance on fast-changing news, we include LiveVQA \cite{fu2025livevqa}. This dataset draws content from major international outlets such as CNN and the BBC and spans 14 categories, including science and sports. It contains \textbf{3,602} pairs generated with GPT-4o, ranging from simple visual checks to complex questions requiring reasoning over accompanying text. This range allows us to test how well models combine visual and textual information in dynamic news environments.

\item \textbf{MAT-Search.} We include MAT-Search \cite{liu2025visual} to evaluate agentic search and multimodal multi-hop reasoning. This benchmark contains \textbf{150} high-quality examples, each manually crafted and verified by human annotators. The questions vary in difficulty and require different depths of reasoning, with more complex queries involving additional inference steps and factual knowledge. These examples are designed to test a model's ability to handle composite problems, retrieve relevant external information, and use tools effectively, providing a focused evaluation of agentic multimodal reasoning.

\end{itemize}

\begin{figure*}[th]
    \centering
    \includegraphics[width=\linewidth]{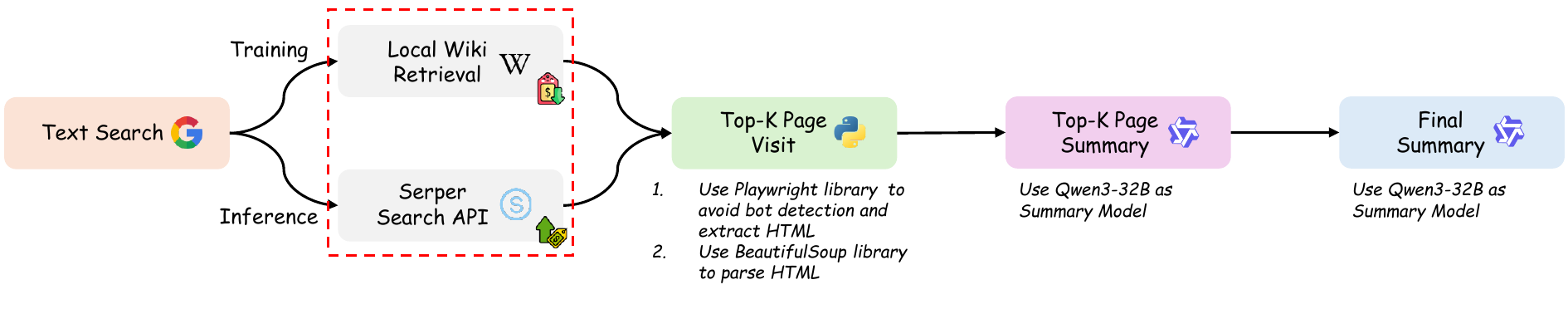}
    \caption{Overview of our Text Search Pipeline. The pipeline utilizes separate retrieval modes for training and inference. Local retrieval from a Wikipedia knowledge base is used during RL training to avoid the prohibitive cost of live web searches, while live web search (via Serper API) is used exclusively during inference. Crucially, the retrieved passages from both separate modes are uniformly processed by a Qwen3-32B summarizer before being passed to the main model.} 
    \label{fig:text_search}
\end{figure*}

\subsection{Visual Reasoning}
For visual reasoning, we mainly evaluate our models on V* Bench \cite{wu2024v}, HR-Bench \cite{wang2025divide} and MME-RealWorld \cite{zhang2024mme}:

\begin{itemize}
    \item \textbf{V* Bench.} V*-Bench~\cite{wu2024v} is designed to evaluate the detailed visual grounding and collaborative reasoning capabilities of VLMs, specifically addressing the need for visual search mechanisms. Sourced from the high-resolution SA-1B dataset, it features images with an average resolution of $2246 \times 1582$ that are visually crowded or contain small details. For this study, we utilize all \textbf{191} images from the benchmark. The dataset includes human-annotated tasks focused on attribute recognition and spatial relationship reasoning, which are deliberately crafted to be unsolvable without precise visual processing and the ability to focus on specific, often obscure, visual elements.
    \item \textbf{HR-Bench.} HR-Bench~\cite{wang2025divide} serves as a specialized benchmark for assessing model performance on ultra-high-resolution inputs, challenging VLMs to overcome the information loss typically associated with image resizing. It evaluates perception capabilities across fine-grained single-instance and cross-instance tasks. To rigorously test the model's scalability and detail preservation, we employ two distinct splits from this dataset: the 4K resolution split and the 8K resolution split, containing \textbf{800} images each. This setup allows for a focused evaluation of the model's stability and accuracy when processing inputs with extreme pixel counts.
    \item \textbf{MME-RealWorld.} MME-RealWorld~\cite{zhang2024mme} is a large-scale, manually annotated benchmark targeting real-world scenarios that are perceptually challenging even for humans. It covers 43 subtasks across five key domains: OCR in the wild, remote sensing, diagrams and tables, video monitoring, and autonomous driving. The images feature high resolutions (average $2000 \times 1500$) and complex, clutter-heavy scenes requiring zooming and multi-step reasoning. In our experiments, we utilize \textbf{23,599} QA samples to evaluate the model's robustness in handling diverse, high-difficulty visual perception tasks in practical applications.
\end{itemize}

\section{Additional Implementation Details}
In this section, we provide more details on the reward model and search pipeline, the differences between the direct answer, RAG, and agentic workflows, and the evaluation metrics used throughout our training and evaluation processes.

\subsection{Reward model}\label{sec:reward_model}
During training for RL, we utilize \textit{Qwen2.5-VL-72B-Instruct} as the LLM-as-a-Judge for all experiments. The judge's response is generated using a greedy setting with a temperature of 0.0. The full prompt is given in Fig.~\ref{fig:judge_prompt}.

\subsection{Comparison between Direct Answer, RAG and Agentic Workflows}\label{sec:workflow}
As described in Section 4.1 of the main paper, we evaluate our models against other baselines using three separate workflows, specifically Direct Answer, RAG, and Agentic Workflows. These workflows are distinguished primarily by the prompts used during inference, which control the tools available to the models. These different prompts are given in Fig.~\ref{fig:agentic_prompt}, Fig.~\ref{fig:direct_prompt} and Fig.~\ref{fig:rag_prompt}.

\subsection{Text search pipeline}\label{sec:search_pipeline}

The structure of our text search pipeline is illustrated in Fig.~\ref{fig:text_search}. Most aspects of the text-search setup are shared between training and inference. The main exception is that the text search method uses local retrieval during training and a live web search during inference. During RL training, we avoid the prohibitive cost of live web searches by using a locally hosted Wikipedia knowledge base built from the 20250901 dump file (\textit{enwiki-20250901-pages-articles.xml.bz2}). Retrieval is performed with the E5-retriever \cite{jin2025search}, and the total number of returned passages is fixed at 5. 

Unlike prior work, such as MMSearch-R1 \cite{wu2025mmsearch}, we do not use the Jina API to extract clean, LLM-friendly text. Instead, to mitigate bot detection, our system uses Playwright to fetch the HTML content. To keep the pipeline simple and efficient, we skip JavaScript rendering and parse the static HTML directly using Python’s BeautifulSoup library. In contrast, during the inference phase, the text-search tool sends its queries through the Serper Search API. Crucially, in both the training and inference phases, the top 5 retrieved passages are first summarized individually by Qwen3-32B \cite{yang2025qwen3}. Following the individual summaries, a final, holistic summary of all 5 passages is then generated. This shared two-stage summarization process ensures that the model learns the core tool-use behaviors on data formatted identically to what it will encounter in a live setting. The complete prompts used for both page and final summarization are identical and are provided in Fig.~\ref{fig:summary_prompt}.

This training design has clear benefits. It is fast and lightweight. The trade-off is that the system cannot read webpages that depend on JavaScript. While supporting these pages could improve performance, this limitation is acceptable because the model still learns the core tool-use behaviors. We observe that these behaviors transfer reliably to inference. Notably, this transfer succeeds even though the model never encounters real Serper Search outputs during its training. 

\subsection{Evaluation Metrics}
We run all evaluations with a sampling temperature of $0.0$ and choose different metrics based on the type of output in each domain. For agentic search tasks, we use an LLM-as-a-Judge setup because the answers are open-ended and require flexible semantic evaluation. In this setup, GPT-4o is used to score the Pass@1 accuracy by comparing the final response with the ground truth. For visual reasoning benchmarks, such as V* Bench~\cite{wu2024v}, HR-Bench~\cite{wang2025divide}, and MME-RealWorld~\cite{zhang2024mme}, we use Exact Match~\cite{lai2025mini} because these datasets mainly contain closed-ended multiple-choice questions that require an exact string match with the correct option. To reduce variance caused by the sampling temperature in these strict visual tasks, we report Avg@8 accuracy for V\* Bench and HR-Bench by averaging the Exact Match score across eight independent attempts for each question. For the large-scale MME-RealWorld benchmark, which has less variance, we report Pass@1 accuracy.

\section{Additional Exprimental Analysis}
\begin{figure}[h]
    \centering
    \includegraphics[width=0.6\linewidth]{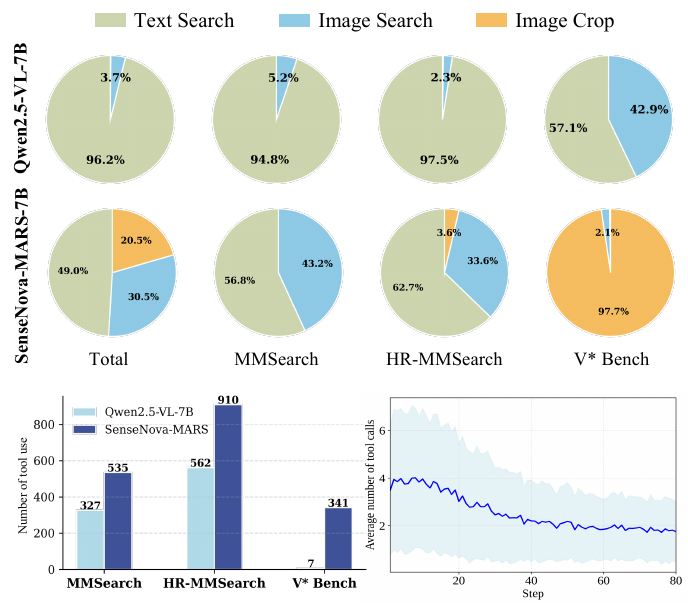}
    \vspace{-5pt}
    \caption{Analysis of tool use behavior. Top: Distribution of tool calls across different benchmarks. Bottom Left: The tool use number in different benchmarks. Bottom Right: Evolution of tool call frequency in the RL training process, indicating that \methodname{} learns more efficient tool invocation strategies.}  
    \label{fig: tool_use}
\end{figure}

In this section, we provide more detailed analysis for the tool use behavior based on \methodname{}-7B.
As shown in Fig.~\ref{fig: tool_use},  The base model Qwen2.5-VL-7B shows an extreme bias. It relies almost exclusively on text search and ignores the image crop tool, making it ineffective for fine-grained perception. 
In contrast, \methodname{}-7B shows a strong multi tool ability that adapts well to task demands. 
\methodname{}-7B illustrates its adaptability by identifying and executing the optimal strategy for each benchmark. For the fine-grained perception tasks in V* Bench, it relies almost entirely on the image crop tool. In the search-oriented MMSearch benchmark, on the other hand, it uses only search tools. In contrast, for the more complex HR-MMSearch, it adopts a hybrid tool-use strategy.
After the cold start, the model's tool-use behavior tends to be redundant. However, RL training progressively streamlines this process, reducing the average tool calls from \textasciitilde{}$4$ to \textasciitilde{}$2$ as shown in Fig.~\ref{fig: tool_use}. This demonstrates that our RL method successfully eliminates superfluous actions, improving both the agent's efficacy and efficiency.

\section{Case Study}
\begin{figure*}[ht]
    \centering
    \includegraphics[width=0.6\linewidth]{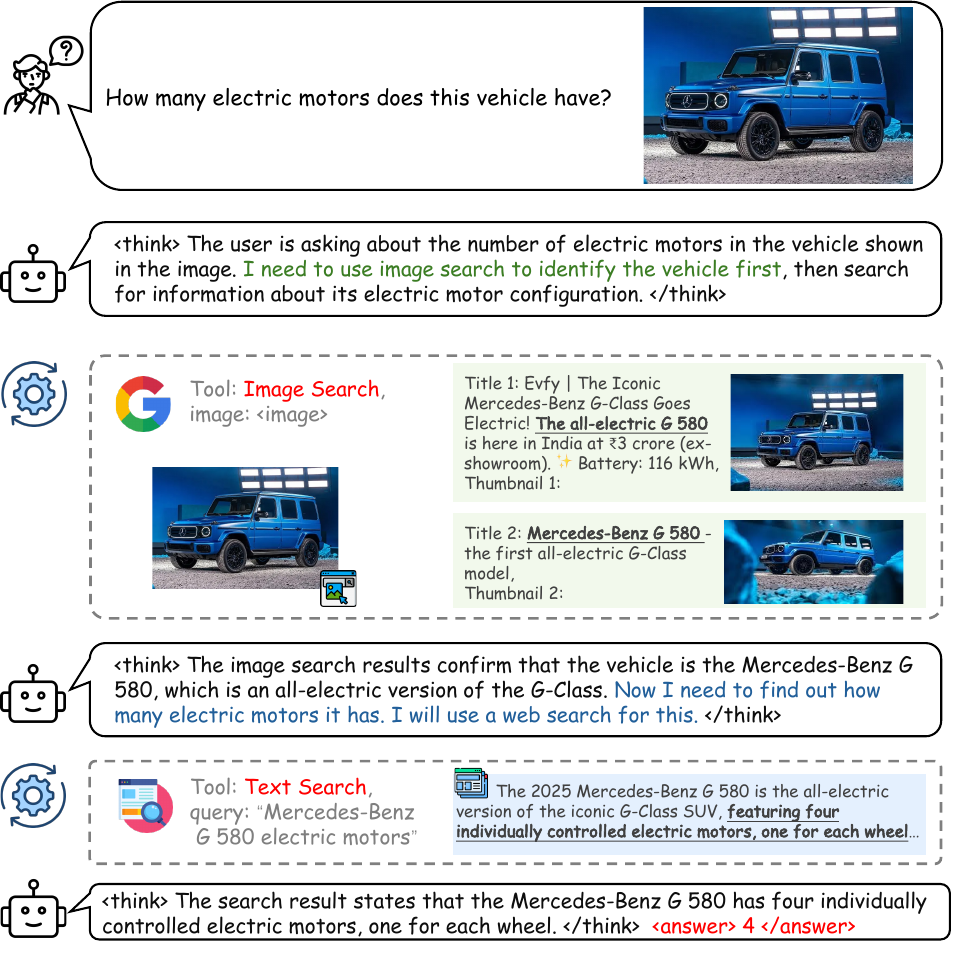}
    \caption{Case study 1 of \methodname.} 
    \label{fig:search_case}
\end{figure*}

\begin{figure*}[ht]
    \centering
    \includegraphics[width=0.6\linewidth]{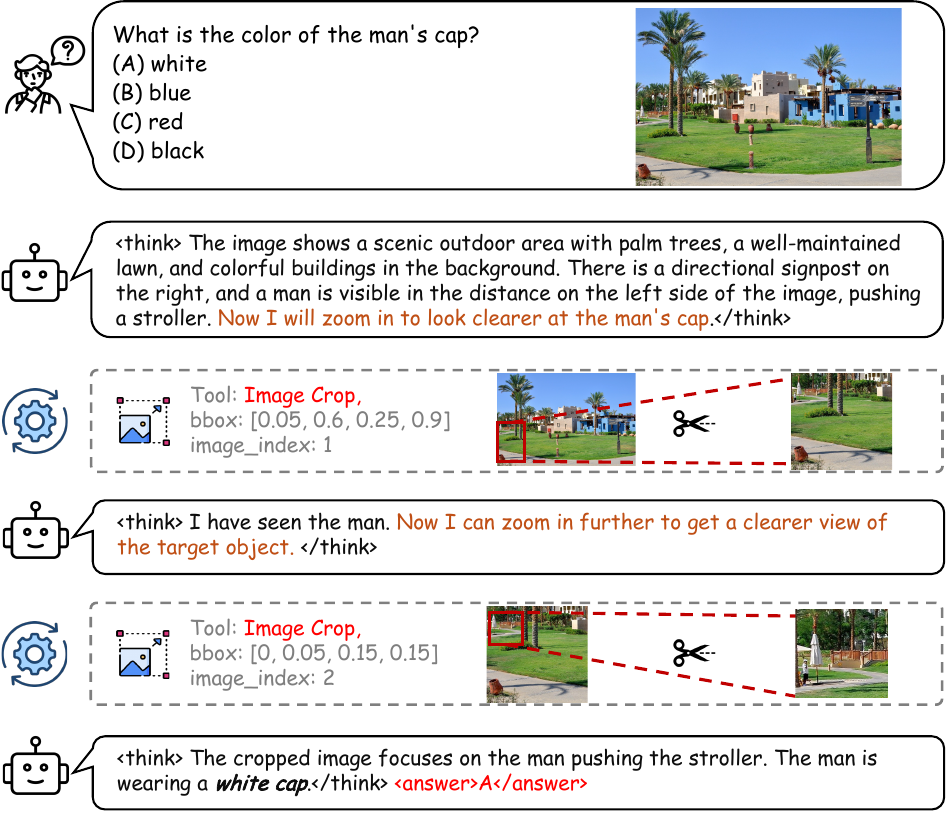}
    \caption{Case study 2 of \methodname.} 
    \label{fig:crop_case}
\end{figure*}

\begin{figure*}[ht]
    \centering
    \includegraphics[width=0.75\linewidth]{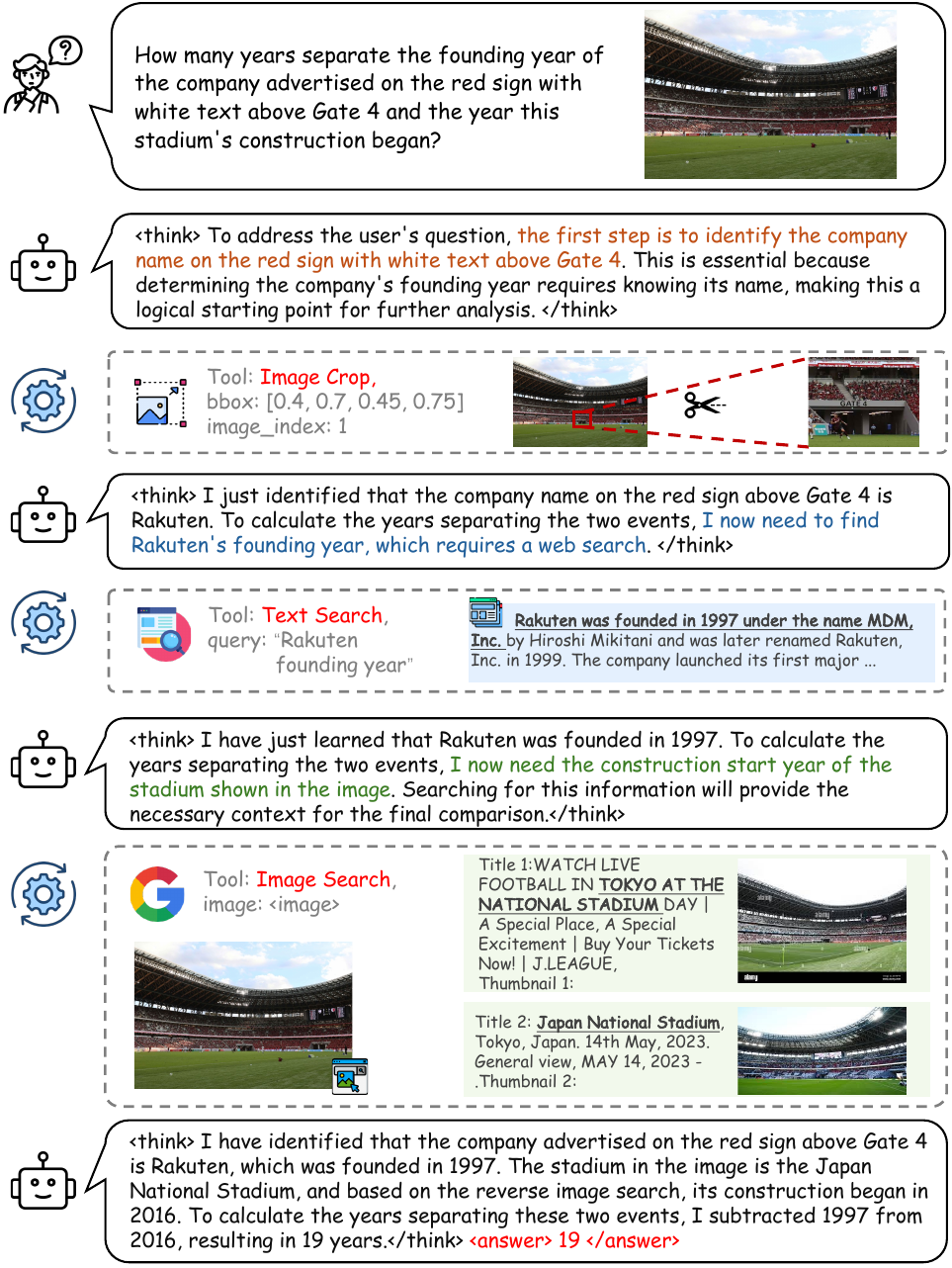}
    \caption{Case study 3 of \methodname.} 
    \label{fig:hrsearch_case}
\end{figure*}

We present more \methodname{} inference cases in Fig.~\ref{fig:search_case}, Fig.~\ref{fig:crop_case} and Fig.~\ref{fig:hrsearch_case}.

\section{Limitations}

\begin{figure*}[ht]
    \centering
    \includegraphics[width=0.7\linewidth]{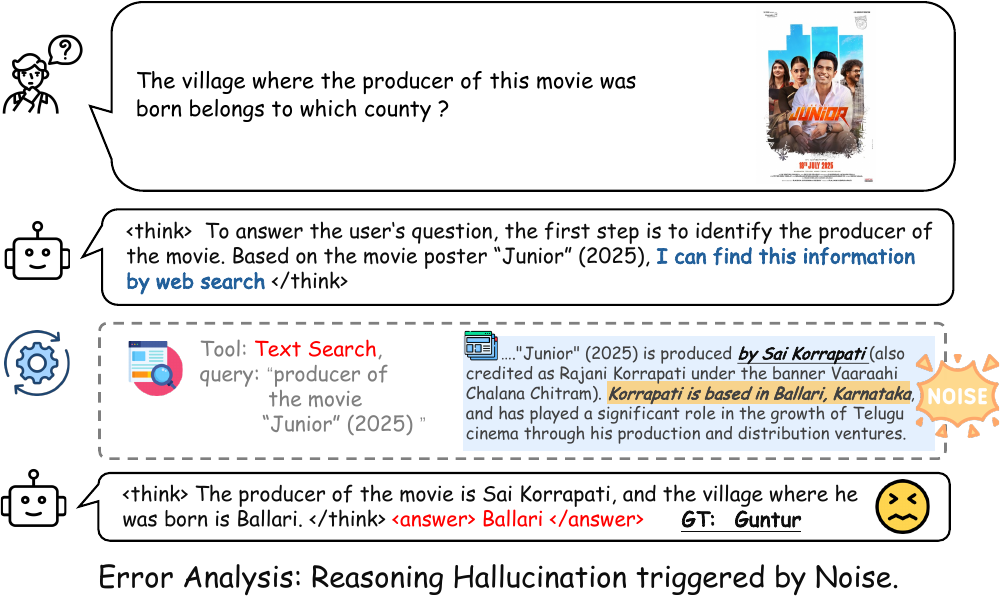}
    \caption{Case study 4 of \methodname.} 
    \label{fig:failure_case1_noise}
\end{figure*}

\begin{figure*}[ht]
    \centering
    \includegraphics[width=0.7\linewidth]{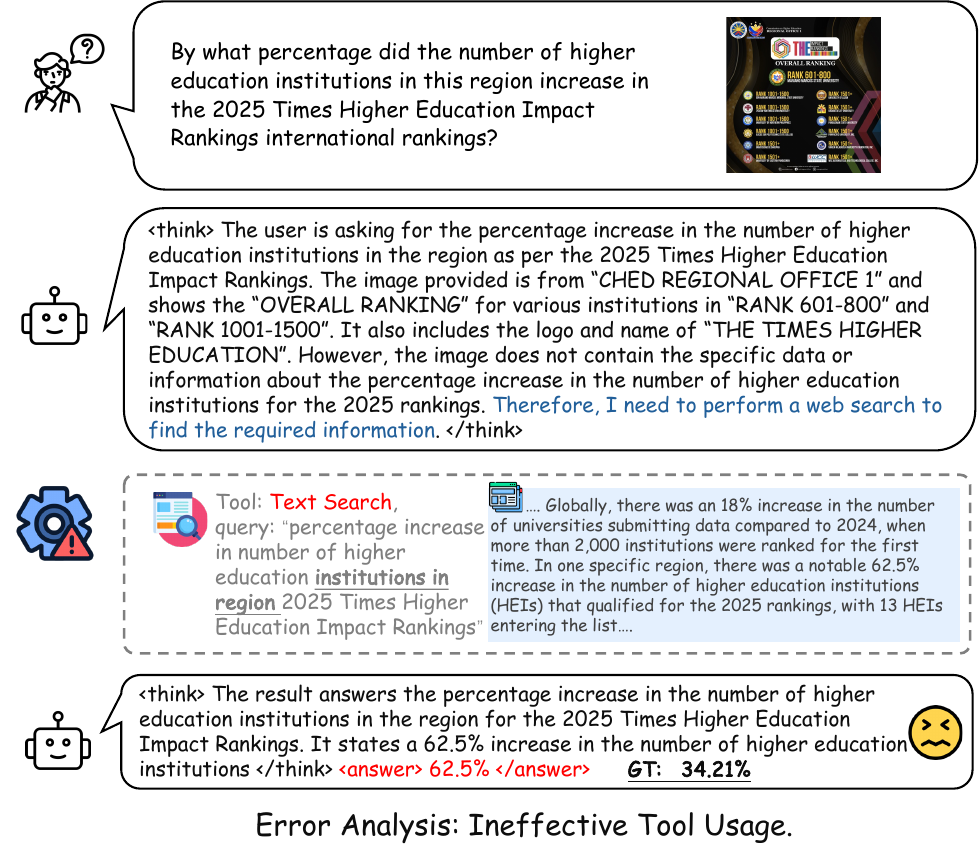}
    \caption{Case study 5 of \methodname.} 
    \label{fig:failure_case2_tool}
\end{figure*}

Despite the strong performance of \methodname{}, our error analysis reveals some limitations:

\begin{itemize}
    \item \textbf{Vulnerability to Retrieval Noise.} 
    As seen in Fig.~\ref{fig:failure_case1_noise}, \methodname{} occasionally fails to distinguish between semantically similar but distinct attributes (e.g., conflating ``based in'' with ``born in'') within retrieved snippets. This suggests that the current reasoning module lacks sufficient robustness against distractor information in open-world search results, leading to hallucinated reasoning paths.
    \item \textbf{Ineffective Tool Usage.} 
    In some scenarios requiring fine-grained visual extraction as shown in Fig.~\ref{fig:failure_case2_tool}, \methodname may fail to ground specific visual entities (e.g., ``CHED Regional Office 1'') into the search query. Instead, it resorts to generic terms (e.g., ``in this region''), resulting in the retrieval of irrelevant global statistics. This indicates a gap in cross-modal alignment during the tool parameter generation phase.
\end{itemize}



\begin{figure*}[t]
\centering

\begin{subfigure}{\textwidth}
\begin{tcolorbox}[
    boxrule=1pt,
    boxsep=2pt,
    colback=gray!20,
    fontupper=\scriptsize,
    fonttitle=\scriptsize\bfseries,
    title=System Message
]

\setlist[itemize]{leftmargin=*, nosep, after=\vspace{4pt}} 

You are an AI assistant tasked with evaluating the correctness of model responses based on an image, question, and ground truth answer. Your judgment should follow these principles:

\medskip

\begin{enumerate}
    \item Consider the image, question, and ground truth answer holistically before evaluating the model's response.
    \item Your decision should be strictly Yes or No, based on whether the model's response is factually accurate and aligns with the ground truth answer.
    \item If the model response is a more specific form of the ground truth answer, it is correct.
    \item If the model response includes all key information but adds minor details, it is correct as long as the extra details are factually correct.
    \item If the model response contradicts, modifies, or omits critical parts of the answer, it is incorrect.
    \item For numerical values, ensure correctness even when presented in different units.
    \item For names, check for first and last name correctness. If the middle name is extra but correct, consider it correct.
    \item For yes/no questions, the response must exactly match "Yes" or "No" to be correct.
    \item If the judgment can be made based solely on the text, you may choose to ignore the input image, as some images may be unfamiliar to you and could affect your judgment. Refer to the image only when necessary to minimize misjudgment.
    \item If there are multiple candidate answers, you can also evaluate the model's response against all of them. If the response aligns with at least one candidate according to the rules above, it should be considered correct.
    \item For multiple choice questions (A, B, C, D), be more lenient. If the model provides the correct letter choice, even with additional text or formatting, consider it correct.
    \item If the model's answer contains the correct choice letter (A, B, C, or D) anywhere in the response, and it's clear this is the intended answer, mark it as correct.
    \item Ignore formatting issues like extra parentheses, brackets, or minor text variations as long as the core answer is correct.
\end{enumerate}

\medskip

\noindent Your output must be in the following format:\\
<judge>Yes/No</judge>\\
<reason>Explanation of why the answer is correct or incorrect.</reason>

\end{tcolorbox}
\end{subfigure}
\hfill

\begin{subfigure}{\textwidth}
\begin{tcolorbox}[
    boxrule=1pt,
    boxsep=2pt,
    colback=gray!20,
    fontupper=\scriptsize,
    fonttitle=\scriptsize\bfseries,
    title=Prompt
]

\textbf{\# Prompt:}

\medskip

Image, Question, and Model Response Evaluation\\
Question: \{\texttt{question}\}\\
Ground Truth Answer: \{\texttt{ground\_truth\_answer}\}\\
Model Response: \{\texttt{model\_response}\}

\medskip

Evaluation Instructions\\
Evaluate whether the Model Response is correct based on the Image, Question and Ground Truth Answer. Follow the predefined judgment rules and provide a clear Yes/No answer along with a justification.

\medskip

Output Format\\
<judge>Yes/No</judge>\\
<reason>Detailed reasoning following the evaluation principles.</reason>

\end{tcolorbox}
\end{subfigure}

\caption{Full prompt used for \textit{Qwen2.5-VL-72B-Instruct} as the LLM-as-a-Judge.}
\label{fig:judge_prompt}
\end{figure*}

\begin{figure*}[t]
\centering

\begin{subfigure}{\textwidth}
\begin{tcolorbox}[
    boxrule=1pt,
    boxsep=2pt,
    colback=gray!20,
    fontupper=\scriptsize,
    fonttitle=\scriptsize\bfseries,
    title=System Message
]

\setlist[itemize]{leftmargin=*, nosep, after=\vspace{4pt}} 

\#Role\\
You are a step-by-step reasoning assistant.\\
Given a question, your task is to solve the problem **one substep at a time**.

\medskip

\#\# Guiding Principles\\
At each turn, you must **either**:
\begin{enumerate}
\item Issue **one specific tool** enclosed in <tool\_call> </tool\_call> tags,  
\item Or provide the **final answer** enclosed in <answer> </answer> tags.  
\end{enumerate}

\medskip

All outputs **must begin with a thought** enclosed in <think> </think> tags, explaining your current reasoning and what to do next.  

\medskip

\#\# Output Format (strict):\\  
Always start with <think>. Do not output the previous reasoning chain. Then, depending on the case, output one of the following:

\medskip

1. If reasoning continues:\\  
<think> Your current reasoning and next plan </think>\\  
<tool\_call> One precise, tool call to assist your reasoning </tool\_call>

\medskip

2. If ready to conclude:\\  
<think> Summarize all reasoning and derive the answer </think>\\  
<answer> Final answer </answer>

\medskip

\# Tools

\medskip

You may call one or more functions to assist with the user query.

\medskip

You are provided with function signatures within <tools></tools> XML tags:\\
<tools>

\medskip

\{"type": "function", "function": \{"name": "web\_search", "description": "Search the web for information you don't have or to verify facts.", "parameters": \{"type": "object", "properties": \{"query": \{"type": "string", "description": "Query to search the web"\}\}, "required": ["query"]\}\}\}\\
\{"type": "function", "function": \{"name": "crop\_image", "description": "Crop the image based on the bounding box coordinates to zoom in on specific regions for detailed analysis.", "parameters": \{ "type": "object", "properties": \{ "bbox": \{ "type": "array", "items": \{ "type": "number" \}, "minItems": 4, "maxItems": 4, "description": "Normalized bounding box [x1, y1, x2, y2], where 0.0 <= x1 < x2 <= 1.0 and 0.0 <= y1 < y2 <= 1.0. (x1,y1) is top-left corner, (x2,y2) is bottom-right corner." \}, "image\_index": \{ "type": "integer", "minimum": 1, "description": "Index of the image to crop: 1 for original input image, 2 for first cropped image, 3 for second cropped image, etc." \}\}, "required": ["bbox", "image\_index"]\}\}\}\\
\{"type": "function", "function": \{"name": "image\_search", "description": "Reverse search the current image to get more information. This function does not accept any text queries or arguments.", "parameters": \{"type": "object", "properties": \{\}\}\}\}

\medskip

</tools>

\medskip

For each function call, return a json object with function name and arguments within <tool\_call></tool\_call> XML tags:\\
<tool\_call>\\
\{"name": <function-name>, "arguments": <args-json-object>\}\\
</tool\_call>

\end{tcolorbox}
\end{subfigure}
\hfill

\begin{subfigure}{\textwidth}
\begin{tcolorbox}[
    boxrule=1pt,
    boxsep=2pt,
    colback=gray!20,
    fontupper=\scriptsize,
    fonttitle=\scriptsize\bfseries,
    title=Prompt
]

\textbf{\# Prompt:}

\medskip

\{\texttt{image}\}\\
\{\texttt{question}\}

\end{tcolorbox}
\end{subfigure}

\caption{Full prompt used during training and inference for the Agentic Workflow.}
\label{fig:agentic_prompt}
\end{figure*}

\begin{figure*}[t]
\centering

\begin{subfigure}{\textwidth}
\begin{tcolorbox}[
    boxrule=1pt,
    boxsep=2pt,
    colback=gray!20,
    fontupper=\scriptsize,
    fonttitle=\scriptsize\bfseries,
    title=System Message
]

\setlist[itemize]{leftmargin=*, nosep, after=\vspace{4pt}} 

You are a helpful assistant.

\end{tcolorbox}
\end{subfigure}
\hfill

\begin{subfigure}{\textwidth}
\begin{tcolorbox}[
    boxrule=1pt,
    boxsep=2pt,
    colback=gray!20,
    fontupper=\scriptsize,
    fonttitle=\scriptsize\bfseries,
    title=Prompt
]

\textbf{\# Prompt:}

\medskip

\{\texttt{image}\}\\
\{\texttt{question}\}

\end{tcolorbox}
\end{subfigure}

\caption{Full prompt used for the Direct Answer workflow.}
\label{fig:direct_prompt}
\end{figure*}

\begin{figure*}[t]
\centering

\begin{subfigure}{\textwidth}
\begin{tcolorbox}[
    boxrule=1pt,
    boxsep=2pt,
    colback=gray!20,
    fontupper=\scriptsize,
    fonttitle=\scriptsize\bfseries,
    title=System Message
]

\setlist[itemize]{leftmargin=*, nosep, after=\vspace{4pt}} 

\#Role\\
You are a step-by-step reasoning assistant.\\
Given a question, your task is to solve the problem **one substep at a time**.

\medskip

\#\# Guiding Principles\\
At each turn, you must **either**:
\begin{enumerate}
\item Issue **one specific tool** enclosed in <tool\_call> </tool\_call> tags,  
\item Or provide the **final answer** enclosed in <answer> </answer> tags.  
\end{enumerate}

\medskip

All outputs **must begin with a thought** enclosed in <think> </think> tags, explaining your current reasoning and what to do next.  

\medskip

\#\# Output Format (strict):\\  
Always start with <think>. Do not output the previous reasoning chain. Then, depending on the case, output one of the following:

\medskip

1. If reasoning continues:\\  
<think> Your current reasoning and next plan </think>\\  
<tool\_call> One precise, tool call to assist your reasoning </tool\_call>

\medskip

2. If ready to conclude:\\  
<think> Summarize all reasoning and derive the answer </think>\\  
<answer> Final answer </answer>

\medskip

\# Tools

\medskip

You may call one or more functions to assist with the user query.

\medskip

You are provided with function signatures within <tools></tools> XML tags:\\
<tools>

\medskip

\{"type": "function", "function": \{"name": "web\_search", "description": "Search the web for information you don't have or to verify facts.", "parameters": \{"type": "object", "properties": \{"query": \{"type": "string", "description": "Query to search the web"\}\}, "required": ["query"]\}\}\}\\
\{"type": "function", "function": \{"name": "crop\_image", "description": "Crop the image based on the bounding box coordinates to zoom in on specific regions for detailed analysis.", "parameters": \{ "type": "object", "properties": \{ "bbox": \{ "type": "array", "items": \{ "type": "number" \}, "minItems": 4, "maxItems": 4, "description": "Normalized bounding box [x1, y1, x2, y2], where 0.0 <= x1 < x2 <= 1.0 and 0.0 <= y1 < y2 <= 1.0. (x1,y1) is top-left corner, (x2,y2) is bottom-right corner." \}, "image\_index": \{ "type": "integer", "minimum": 1, "description": "Index of the image to crop: 1 for original input image, 2 for first cropped image, 3 for second cropped image, etc." \}\}, "required": ["bbox", "image\_index"]\}\}\}

\medskip

</tools>

\medskip

For each function call, return a json object with function name and arguments within <tool\_call></tool\_call> XML tags:\\
<tool\_call>\\
\{"name": <function-name>, "arguments": <args-json-object>\}\\
</tool\_call>

\end{tcolorbox}
\end{subfigure}
\hfill

\begin{subfigure}{\textwidth}
\begin{tcolorbox}[
    boxrule=1pt,
    boxsep=2pt,
    colback=gray!20,
    fontupper=\scriptsize,
    fonttitle=\scriptsize\bfseries,
    title=Prompt
]

\textbf{\# Prompt:}

\medskip

\{\texttt{image}\}\\
\{\texttt{question}\}

\medskip

To help you answer the question, here are reverse image search results for the given image.

\medskip

Reverse image search results:

\medskip

\{\texttt{image\_search\_results}\}

\end{tcolorbox}
\end{subfigure}

\caption{Full prompt used for the RAG Workflow.}
\label{fig:rag_prompt}
\end{figure*}

\begin{figure*}[t]
\centering

\begin{subfigure}{\textwidth}
\begin{tcolorbox}[
    boxrule=1pt,
    boxsep=2pt,
    colback=gray!20,
    fontupper=\scriptsize,
    fonttitle=\scriptsize\bfseries,
    title=System Message
]

\setlist[itemize]{leftmargin=*, nosep, after=\vspace{4pt}} 

You are a helpful assistant. Your task is to summarize the main content of the given web page in no more than five sentences. Your summary should cover the overall key points of the page, not just parts related to the user's question.

\medskip

If any part of the content is helpful for answering the user's question, be sure to include it clearly in the summary. Do not ignore relevant information, but also make sure the general structure and main ideas of the page are preserved. Your summary should be concise, factual, and informative.

\end{tcolorbox}
\end{subfigure}
\hfill

\begin{subfigure}{\textwidth}
\begin{tcolorbox}[
    boxrule=1pt,
    boxsep=2pt,
    colback=gray!20,
    fontupper=\scriptsize,
    fonttitle=\scriptsize\bfseries,
    title=Prompt
]

\textbf{\# Prompt:}

\medskip

Webpage Content (first 30000 characters) is: \{\texttt{content}\}\\
Question: \{\texttt{question}\}

\end{tcolorbox}
\end{subfigure}

\caption{Full prompt used by Qwen3-32B to perform page summary and final summary.}
\label{fig:summary_prompt}
\end{figure*}

\end{document}